\documentclass{article} % For LaTeX2e
\usepackage{iclr2025_conference,times}

\usepackage{amsmath,amsfonts,bm}

\def\eqref#1{equation~\ref{#1}}
\def\1{\bm{1}}

\DeclareMathAlphabet{\mathsfit}{\encodingdefault}{\sfdefault}{m}{sl}
\SetMathAlphabet{\mathsfit}{bold}{\encodingdefault}{\sfdefault}{bx}{n}

\DeclareMathOperator*{\argmin}{arg\,min}

\usepackage{hyperref}
\usepackage{url}
\usepackage{graphicx}
\usepackage{subcaption}
\usepackage{caption}
\usepackage{float}
\usepackage{booktabs}
\usepackage{array}
\usepackage[T1]{fontenc}

\title{Interpretable Unsupervised Joint Denoising and Enhancement for Real-World low-light Scenarios}

\author{Huaqiu Li, Xiaowan Hu, Haoqian Wang \thanks{Corresponding author.} \\
Tsinghua Shenzhen International Graduate School\\
Tsinghua University\\
\texttt{lihq23@mails.tsinghua.edu.cn}
}

\iclrfinalcopy
\begin{document}

\maketitle
\vspace{-4mm}
\begin{abstract}
Real-world low-light images often suffer from complex degradations such as local overexposure, low brightness, noise, and uneven illumination. Supervised methods tend to overfit to specific scenarios, while unsupervised methods, though better at generalization, struggle to model these degradations due to the lack of reference images. To address this issue, we propose an interpretable, zero-reference joint denoising and low-light enhancement framework tailored for real-world scenarios. Our method derives a training strategy based on paired sub-images with varying illumination and noise levels, grounded in physical imaging principles and retinex theory. Additionally, we leverage the Discrete Cosine Transform (DCT) to perform frequency domain decomposition in the sRGB space, and introduce an implicit-guided hybrid representation strategy that effectively separates intricate compounded degradations. In the backbone network design, we develop retinal decomposition network guided by implicit degradation representation mechanisms. Extensive experiments demonstrate the superiority of our method. Code will be available at \url{https://github.com/huaqlili/unsupervised-light-enhance-ICLR2025}.
\end{abstract}
\vspace{-4mm}
\section{introduction}
\vspace{-3mm}
Low-light image enhancement is a significant research area in computer vision and image processing. The inherently low signal-to-noise ratio of such images can adversely impact downstream tasks, such as object detection~\cite{objectdetection}, image segmentation~\cite{segmetation}, and face recognition~\cite{lightface}. Moreover, the widespread application of low-light enhancement in fields like nighttime photography~\cite{nightenhance,apsf}, astronomical observation~\cite{tianwen}, and autonomous driving~\cite{lightdriving} underscores its critical importance in low-level vision tasks.

Real-world low-light enhancement presents numerous challenges, requiring simultaneous handling of issues such as brightness, contrast, artifacts, and noise. Over the past few decades, traditional methods like gamma correction~\cite{gamma}, histogram equalization~\cite{histogram}, and retinex theory~\cite{retinex} have been developed. However, these methods focus on single-dimensional brightness issues and struggle with complex real-world scenes, while their handcrafted priors often lack generalization for diverse conditions. 

In recent years, learning-based methods for low-light enhancement have achieved significant progress. However, these approaches often rely on paired (e.g.~\cite{kind,cai2023retinexformer,wu2022uretinex,retinex-diffusion,retinexmamba}) or unpaired (e.g.~\cite{jiang2021enlightengan,nerco}) data, making it challenging to collect large-scale datasets. Additionally, discrepancies in brightness between reference images can disrupt model fitting, making the development of efficient zero-reference methods crucial.

\begin{figure}[t] % 'h' 表示在此处插入图片
    \centering % 居中
    \includegraphics[width=\textwidth]{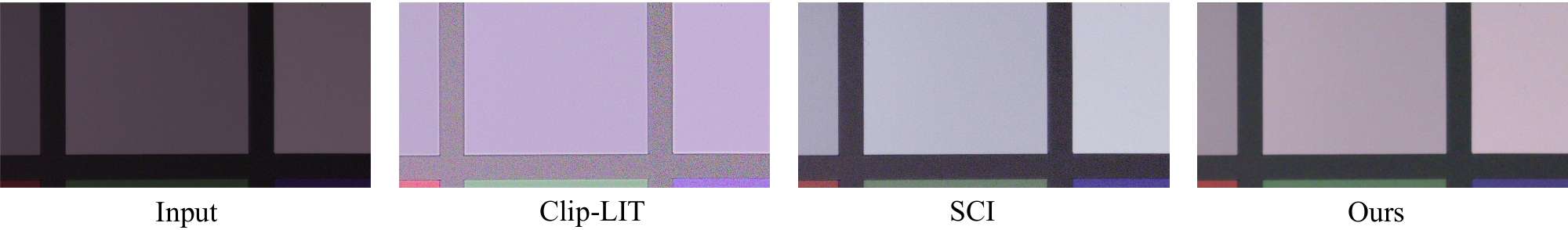} % 调整图片宽度和路径
    \vspace{-4mm}
    \caption{Compared with state-of-the-art methods~\cite{clip,SCI} on the SIDD dataset, our approach achieves the best results in denoising, enhancement, and color fidelity grounded in real-world imaging principles.} % 图片标题
    \vspace{-5mm}
    \label{fig:feasor} % 图片标签，用于引用
\end{figure}

Current zero-reference low-light enhancement methods, such as Zero-DCE~\cite{ZERODCE}, utilize curve learning for iterative optimization, but does not account for noise degradation. Approaches like SCI~\cite{SCI} and RUAS~\cite{RUAS} follow a similar iterative strategy, integrating denoising modules. However, while separate denoising modules are designed for end-to-end training, they rely on specific, lengthy loss functions that lack generalization across various noise patterns. Other methods~\cite{fan2022multiscale} address multiple degradation tasks through multi-stage learning. They often overlook the error accumulation during the optimization process (e.g., noise becomes more complex after low-light enhancement). Furthermore, as Fig.\ref{fig:feasor} shows, these methods generally fail to differentiate feature layers for multiple degradation modes, leading to confusion and ambiguity during the restoration process.

To address the aforementioned challenges, we propose a zero-reference joint denoising and enhancement method grounded in real-world physical models. Specifically, we introduce a self-supervised image denoising method based on neighboring pixel masking, alongside a self-supervised enhancement strategy that combines random gamma adjustment with retinex theory. By obtaining sub-image pairs with varying illumination and noise levels, the framework is capable of tackling the complex degradation issues caused by low-light conditions. Additionally, we employed DCT to model physical priors that reflect various degradations, and designed a global learning-based encoder to extract implicit degradation representations from them. In the backbone network design, we develop retinal decomposition network guided by implicit degradation representation mechanisms. This approach allows us to separate and address complex degradations in the frequency domain, rather than sequentially handling features as in previous methods. Extensive experiments demonstrate that our method offers significant advantages over the current SOTA approaches.

The main contributions of this paper are as follows:
\vspace{-2mm}
\begin{itemize}
    \item By preprocessing the original low-light image to generate paired sub-images with varying illumination and noise levels, followed by retinal decomposition, we derived and validated a physically sound unsupervised joint denoising and enhancement framework.
    \item We utilized DCT to model physical priors that capture intricate compounded degradations, and designed a globally learned encoder to extract implicit degradation representations from these priors.
    \item We developed a hybrid-prior attention transformer network that integrates degradation features to reconstruct the reflection map, while adaptively enhancing the illumination.
    \item Extensive experiments on multiple real-world datasets demonstrate that our method achieves superior performance across several metrics compared to SOTA approaches.
\end{itemize}
\vspace{-4mm}
\section{Related works}
\vspace{-2mm}
\subsection{self-supervised/unsupervised low-light image enhancement}
The development of self-supervised and unsupervised low-light enhancement follows two main approaches: zero-reference and unpaired learning. Zero-DCE~\cite{ZERODCE} introduced a curve-based iterative method for zero-reference enhancement, later refined by Zero-DCE++~\cite{zerodceplus} for better efficiency. Methods like RUAS~\cite{RUAS} and SCI~\cite{SCI} extend this approach with denoising modules for handling complex degradations. However, these approaches often struggle with interpretability and modeling complex degradations. In contrast, unpaired learning leverages low-light and normal-light image pairs from different scenes or varying illumination within the same scene. GAN-based methods like EnlightenGAN~\cite{jiang2021enlightengan} and NeRCo~\cite{nerco} use cyclical networks for bidirectional image transformation learning between domains. PairLIE~\cite{pairlie} processes low-light images with varying degradations from the same scene using retinal theory. Although these methods demonstrate strong generative abilities, their performance can be constrained by inconsistent normal-light references and difficulties in normalizing illumination distributions.

\subsection{Frequency-domain analysis in image processing}
DCTconv~\cite{dctconv} integrates convolution with IDCT to form a novel layer that facilitates network pruning.~\cite{xie2021learning} introduces a frequency-aware dynamic network that leverages DCT in image super-resolution to reduce computational cost. To improve content preservation,~\cite{cai2021frequency} employs a Fourier frequency spectrum consistency constraint for image translation. Recently, frequency domain processing has gained significant attention. \cite{zou2024freqmamba} demonstrates that degradation predominantly affects amplitude spectra, while FSI~\cite{liu2023fsi} designs a frequency-spatial interactive network to address under-display camera image restoration.~\cite{zou2022joint} employs wavelet transforms to disentangle frequency domain information, using a multi-branch network to recover high-frequency details. WINNet~\cite{ou2024winnet} combines wavelet-based and learning-based methods to construct a reversible, interpretable network with strong generalization capabilities. FCDiffusion~\cite{fcdiffusion} utilizes DCT to filter feature maps, achieving controlled generation across different frequency bands.

\section{method}
\subsection{theoretical basis}
\subsubsection{Retinex theory}
The traditional Retinex image enhancement algorithm~\cite{retinex,lolv1} simulates human visual perception of brightness and color. It decomposes image \({I} \in \mathbb{R}^{H \times W \times 3}\) into the illumination component \({L} \in \mathbb{R}^{H \times W \times 3}\) and the reflection component \({R} \in \mathbb{R}^{H \times W \times 3}\). This conclusion can be expressed by the following formula:
\begin{equation}
I = R \circ L
\end{equation}
where \(\circ\) denotes element-wise multiplication. The reflection component \(R\) is determined by the intrinsic properties of the object, while the illumination component \(L\) represents the lighting intensity. However, the traditional Retinex algorithm does not account for complex degeneration produced by unbalanced light distribution or real-world dark scenes in low-light conditions, and this loss of quality is further amplified with the enhancement of the image. Therefore, we add the noise disturbance term $N$ on  the reflection component as the basis of theoretical analysis:
\begin{equation}
I = (R+N) \circ L
\end{equation}
In most low-light scenarios, $N$ is modeled as zero-mean Poisson noise.
\subsubsection{neighboring pixel masking in self-supervised denoising}
Image denoising represents a classic ill-posed problem within the domain of image restoration. This signifies the existence of multiple potential solutions for the same noisy scene. Previous image denoising models~\cite{dncnn,denoisingreview} typically require paired input of noisy images $\mathbf{y}_i$ and corresponding clean images $\mathbf{x}_i$ to train the network effectively.
\begin{equation}
\argmin_{\theta}\sum_{i}^{}L(f_{\theta } (\mathbf{y}_i),\mathbf{x}_i)
\end{equation}
Here, $\theta$ represents parameters that need to be optimized. However, in practical scenarios, obtaining paired images is often challenging or even impossible. As a result, a series of self-supervised and unsupervised methods have emerged utilizing only noisy images for training.

The theoretical foundation of N2N\cite{n2n} is rooted in point estimation, which estimates the true value of a series of observations \{$\mathbf{x}_1$, $\mathbf{x}_2$, ..., $\mathbf{x}_n$\}. The objective is to find a value $\mathbf{z}$ that minimizes the sum of distances to all the observed values, serving as the estimation. When using $\mathcal{L}_2$ loss for estimation, replacing $x$ with another observation $\mathbf{z}$ having the same mean value does not alter the result. 

Extending this theoretical point estimation framework to training neural network regressors, the optimization objective of the network can be transformed into:
%\vspace{-2.5mm}
\begin{equation}
\label{equ:n2n}
    \argmin_{\theta}\sum_{i}^{}L(f_{\theta } (\mathbf{y}_i),\mathbf{z})
%\vspace{-2.5mm}
\end{equation}
This implies that when training a denoising network, if we replace the clean images $\mathbf{x}_i$ with noisy images $\mathbf{z}$, which have zero-mean noise, the optimization results using L2 loss will be equivalent to those trained by pairs of noisy-clean images. This assumption forms the foundation of our work. 

\begin{figure}[t] % 'h' 表示在此处插入图片
    \centering % 居中
    \includegraphics[width=14cm]{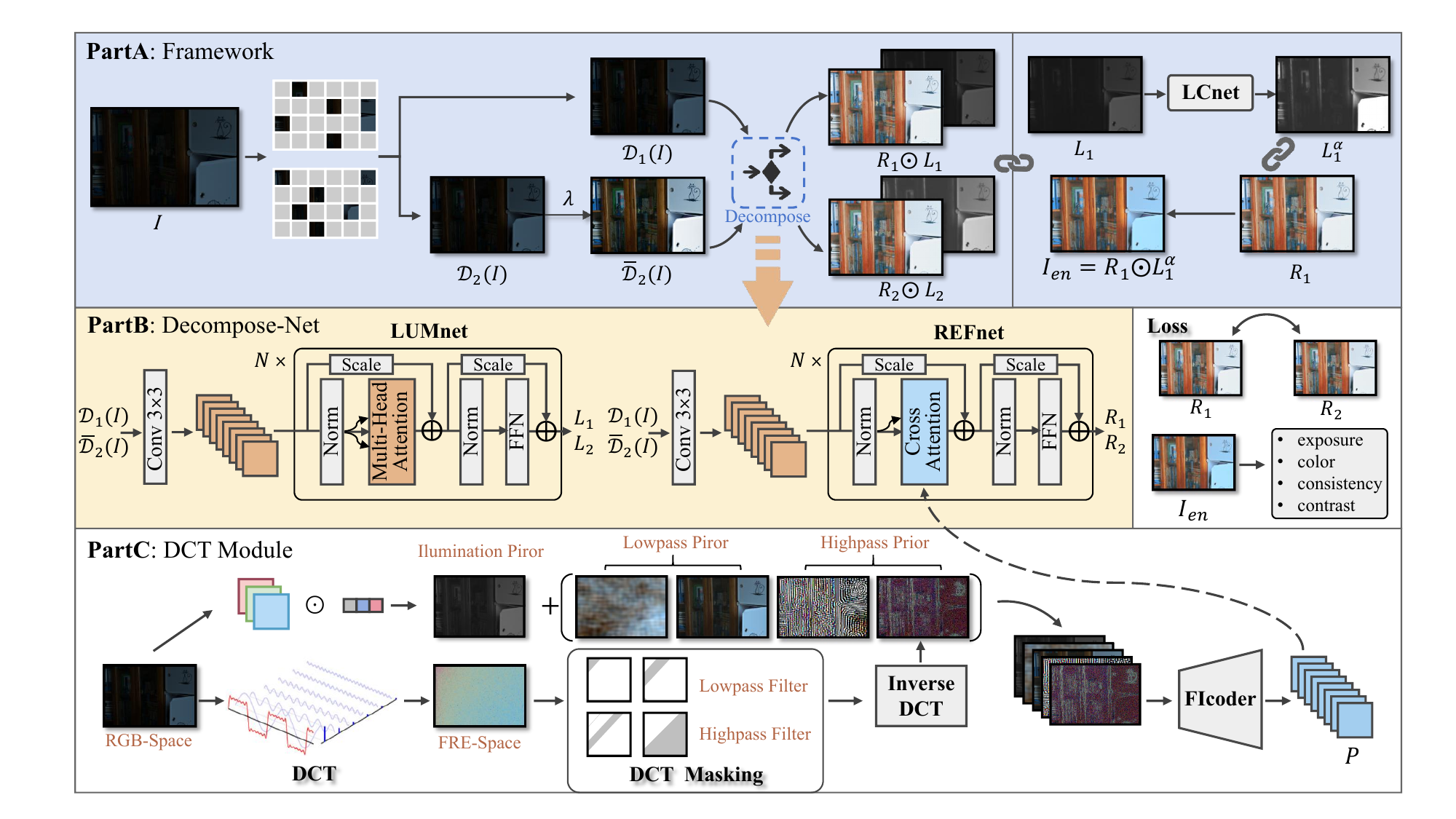} % 调整图片宽度和路径
    \caption{The pipeline of our proposed method: First, we preprocess the low-light full-resolution image \(I\) using pixel masks and gamma-based nonlinear enhancement, generating sub-images with varying illumination and noise levels. These are then processed through Decompose-Net, which uses a transformer architecture integrating hybrid degradation representations, incorporating cross-attention to inject guiding embeddings. Subsequently, LCnet enhances the illumination map.}
    \vspace{-5mm}% 图片标题
    \label{fig:pipe} % 图片标签，用于引用
\end{figure}

\subsection{Overall Architecture}
Building on the aforementioned theoretical foundation, we express a low-light image \( I = (R + N) \circ L \), where \( N \) represents a zero-mean noise distribution. Our objective is to generate images of the same scene with differing noise observations, ensuring that the noise remains zero-mean and the denoised ground truth is consistent across these images. In scenarios where a normal-light reference image is unavailable, we propose to generate two sub-images at 1/4 resolution through a process of neighboring masking \( \mathcal{D} \). Specifically, the original image \( I \) is partitioned into multiple 2x2 pixel patches. From each patch, two adjacent pixels are randomly selected and assigned to corresponding regions in the two sub-images. The resulting sub-images can thus be mathematically formulated as:
\begin{equation}
\mathcal{D}_1(I)=(R_1 + N_1) \circ L_1, \mathcal{D}_2(I)=(R_2 + N_2) \circ L_2
\end{equation}
Here, \( N_1 \) and \( N_2 \) represent noise components that follow a shared distribution, \( R_1 \) and \( R_2 \) are highly similar in pixel values, and \( L_1 \) and \( L_2 \) correspond to the same lighting conditions.

Previous study~\cite{pairlie} has indicated that if images of the same scene under different illumination conditions can be obtained, deep learning can be employed to decompose the corresponding reflectance \( R \), with the principle that the reflectance \( R_1 \) and \( R_2 \) should theoretically be identical. To generate a supervision signal with different illumination, we apply gamma correction to $\mathcal{D}_2(I)$ and get $\overline{\mathcal{D}}_2(I)$. We avoid applying gamma correction directly to the original image $I$ because the noise $N$ would be preserved at nearly the same level, making the network learn an identity mapping. After obtaining the enhanced image $\overline{\mathcal{D}}_2(I)$, and given that \( N_2 \) is relatively small compared to the pixel values, we further perform a Taylor series expansion on it:
\begin{equation}
\overline{\mathcal{D}}_2(I)=\mathcal{D}_2(I)^{\lambda}=(R_2 + N_2)^{\lambda} \circ L_2^{\lambda}\approx (R_2^\lambda+\lambda R_2^{\lambda-1}N_2)\circ L_2^\lambda=(R_2+\lambda N_2)\circ R_2^{\lambda-1}\circ L_2^\lambda
\end{equation}
Here, \(\lambda\) represents the gamma enhancement factor, and \(R^{\lambda-1} \approx 1\) when \(\lambda\) is close to 1. The original equation can thus be further rewritten as \((R_2+\lambda N_2)\circ \overline{L}_2\), \(\overline{L}_2=L_2^\lambda\), leading to the final expressions for the two sub-images:
\begin{equation}
\mathcal{D}_1(I)=(R_1 + N_1) \circ L_1, \overline{\mathcal{D}}_2(I)=(R_2+\lambda N_2)\circ \overline{L}_2
\end{equation}
In this formulation, \(R_1\) and \(R_2\) share the same ground truth reflectance, as they exist within the same scene. Meanwhile, \(N_1\) and \(\lambda N_2\) represent zero-mean noise distributions that are non-identical. Additionally, the first and second images encompass different illumination conditions. Therefore, by simply constraining \((R_1 + N_1)\) and \((R_2+\lambda N_2)\) to be equal, we can construct a self-supervised network jointly performing denoising and enhancement (DEnet).
\begin{figure}[t] % 'h' 表示在此处插入图片
    \centering % 居中
    \includegraphics[width=12cm]{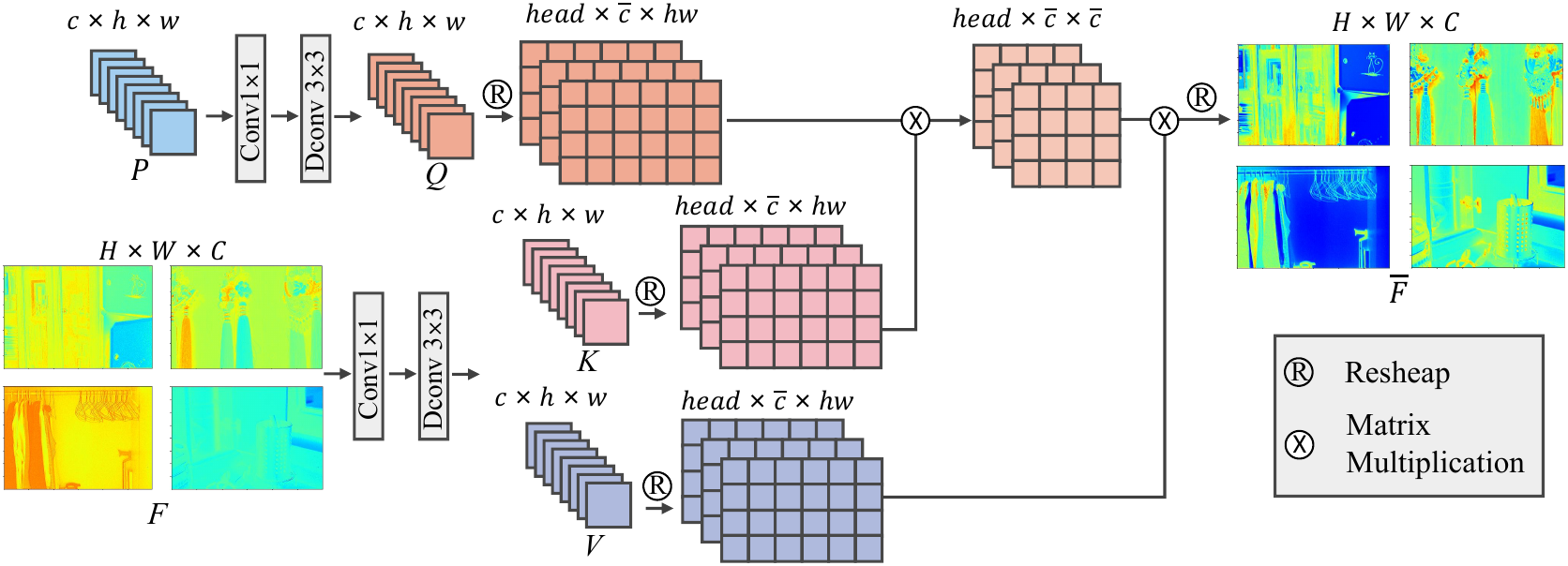} % 调整图片宽度和路径
    \vspace{-2mm}
    \caption{Illustration of the hybrid prior degradation representation guided by multi-head cross attention. After processing, the feature maps exhibit clearer hierarchical structure and reduced noise.}
    \vspace{-5mm}% 图片标题
    \label{fig:atten} % 图片标签，用于引用
\end{figure}

As illustrated in Fig.\ref{fig:pipe}, the overall architecture of DEnet is primarily divided into four components: the Frequency-Illumination representation Encoder (FIcoder), the Reflectance Map Extraction Network (REFnet), the Illumination Map Extraction Network (LUMnet), and the Light Correction Network (LCnet). REFnet and LUMnet are employed to extract the reflectance maps $R_1$, $R_2$, and illumination maps $L_1$, $\overline{L}_2$ from sub-images $\mathcal{D}_1(I)$ and $\overline{\mathcal{D}}_2(I)$. In LUMnet, each transformer block is divided into a self-attention computation module and a gating module. In contrast, REFnet, tasked with reflectance map extraction, requires the degradation representations to perform cross-attention calculations with feature tokens ss illustrated in the Fig.\ref{fig:atten}. LCnet processes its features using a transformer and then applies global average pooling. The pooled features are passed through two linear layers to scale them into a one-dimensional enhancement factor to correct the illumination map, which is subsequently multiplied with the reflectance map to produce the final corrected image \( I_{en} \). 

\subsection{Frequency-illumination prior encoder}
\begin{figure}[t] % 'h' 表示在此处插入图片
    \centering % 居中
    \includegraphics[width=\textwidth]{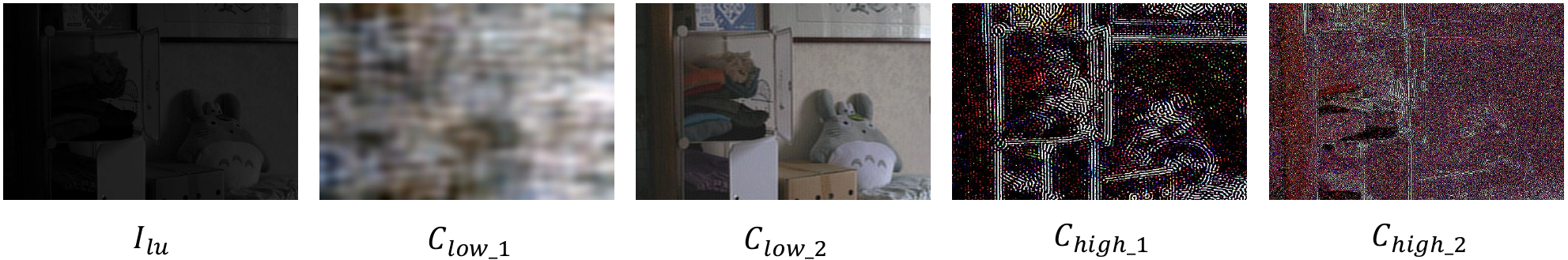} % 调整图片宽度和路径
    \caption{The visualization of the five image priors. They represent chromaticity, semantic information, edge contours, and noise intensity.} % 图片标题
    \vspace{-3mm}
    \label{fig:prior} % 图片标签，用于引用
\end{figure}

FIcoder is primarily designed to obtain degradation representations from illumination and frequency domain priors, which are then integrated with feature maps through cross-attention mechanisms in REFnet. The fusion of multiple priors enhances the model's generalization capability across diverse and complex degradations. As illustrated in the Fig.\ref{fig:prior}, the illumination prior $I_{lu}$ represents the image's luminance information, while the four frequency domain priors $C_{low\_1}, C_{low\_2}, C_{high\_1}, C_{high\_2}$, ranging from low to high frequencies, capture information on chromaticity, semantics, edge contours, and noise intensity, respectively.

First, we extract the illumination prior \(I_{lu} = mean_c(I)\), which is the mean value of the sub-image across the channel dimension, representing the overall brightness level of the image. As for frequency prior, we use channel-wise 2D DCT to convert the spatial-domain image $I$ into the frequency-domain counterpart $F$. Different spectral bands in the DCT domain encode different image visual attributes degradation representation analysis of input images. To obtain the frequency spectrum maps across four frequency bands, we define four masks:
%\begin{equation}
    \begin{align}    
    &M_{low\_1}(u,v)=1\ \ if\ \ u+v\leq t\ \ else\ \ 0, M_{low\_2}(u,v)=1\ \ if\ \ u+v\leq 3t\ \ else\ \ 0,\\
    &M_{high\_1}(u,v)\!=\!1\ \ if\ \ 2t\!<\!u+v\!\leq\!4t\ \ else\ \ 0, M_{high\_2}(u, v)\!=\!1\ \ if\ \ u+v\geq 5t\ \ else\ \ 0.
    \end{align}
%\end{equation}
\begin{equation}
    F^{*}=F\times M^{*},
\end{equation}
where $*$ $\in$ \{low\_1, low\_2, high\_1, high\_2\}, and $t$ represents the manually set bandwidth hyperparameter. We apply the masks $M_{*}$ to the frequency spectrum feature maps $F$ to filter them according to different frequency bands. By performing an inverse Discrete Cosine Transform (IDCT) on these filtered maps $F_{*}$, we obtain the corresponding spatial domain feature images $C_{*}$.

Finally, we combine \( I_{lu} \) $\in \mathbb{R}^{H \times W \times 1}$, $C_{low\_1}, C_{low\_2}, C_{high\_1}$ and $C_{high\_2}$\(\in \mathbb{R}^{H \times W \times 3}\) through a convolutional network-based Illumination-Frequency Prior Encoder. This encoder constructs the implicit representation \( P \)\(\in \mathbb{R}^{H \times W \times C}\), based on separating degradation features. During training, the FIcoder processes the input sub-images \(\mathcal{D}_1(I)\) and \(\overline{\mathcal{D}}_2(I)\), generating the corresponding degradation representations \(P_1\) and \(P_2\).

\subsection{loss function}
During model training, DEnet performs the following computations:
\begin{equation}
    I_{en}=DE(\mathcal{D}_1(I))=R_1\circ L_1^\alpha,R_1=REF(\mathcal{D}_1(I),P_1),L_1=LUM(\mathcal{D}_1(I)),\alpha =LC(L_1)
\end{equation}
During inference, we input the original-resolution low-light image $I$, multiply the decomposed reflection component \( R \) with the corrected illumination \( L \), and obtain the final enhanced result.

The loss function for this method is primarily divided into two aspects: 
\textbf{1) Retinex Decomposition Loss:} This loss constrains the retinex decomposition to ensure that the resulting reflectance and illumination maps are consistent with the underlying physical assumptions.
\textbf{2) Self-supervised Enhancement Loss:} This loss is designed to regulate the enhanced image \(I_{en}\) by imposing constraints on brightness, contrast, saturation, and other factors, ensuring that the enhancement aligns with desired visual qualities.

The Retinal Decomposition Loss we employ is primarily divided into two parts: the first is \( \mathcal{L}_R \), as mentioned earlier, which primarily constrains the \( L_2 \) distance between the reflectance maps $R_1$ and $R_2$ derived from $\mathcal{D}_1(I)$ and $\overline{\mathcal{D}}_2(I)$; the second is \( \mathcal{L}_L \), which imposes smoothness constraints on the illumination maps and ensures that the product of the decomposed maps equals the original image. The expressions for these two losses are shown as follows:
\begin{equation}
    \mathcal{L}_R=\left \| REF(\mathcal{D}_1(I),P_1)-REF(\overline{\mathcal{D}}_2(I),P_2) \right \| ^{2}_2+\omega_{reg}\mathcal{L}_{reg}
\end{equation}
\vspace{-2mm}
\begin{equation}
    \mathcal{L}_L=\left \| R_1\circ L_1-\mathcal{D}_1(I)\right \| ^{2}_2+\left \| L_1-L_0\right \| ^{2}_2+\left \| R_1-\frac{\mathcal{D}_1(I)}{L_1.detach()} \right \| ^{2}_2+\bigtriangledown L_1,L_0=\! \max_{c\in \{r,g,b\}}\!\mathcal{D}_1(I)_c
\end{equation}
Here, \(P_1\) and \(P_2\) represent the degradation representations extracted by the FIcoder from the sub-images \(\mathcal{D}_1(I)\) and \(\overline{\mathcal{D}}_2(I)\), respectively. \(\bigtriangledown L_1\) denotes the gradient of the illumination map. We add a regularization term $\mathcal{L}_{reg}$ to align gradients and test the original-scale images. The masked testing results are compared with sub-image reflectance maps via \(L_2\)-norm, ensuring the consistency of \(R_1\) and \(R_2\) across scales, enhancing generalization and training stability. 
\begin{equation}
    \mathcal{L}_{reg} \!=\! \left \| REF(\mathcal{D}_1(I),P_1)\!-\!REF(\overline{\mathcal{D}}_2(I),P_2)\!-\!(\mathcal{D}_1(REF(I,P))-\overline{\mathcal{D}}_2(REF(I,P)) \right \| ^{2}_2
    \label{equ:reg}
\end{equation}
For the Self-supervised Enhancement Loss, we designed two components: the consistency loss \( L_{con} \) and the enhancement loss \( L_{enh} \):
\begin{equation}
    \mathcal{L}_{con}=\frac{1}{K}\sum_{i=1}^{K}\sum_{j\in \sigma (i)}^{}(\left | I_{en,i}-I_{en,j} \right |-\left | \mathcal{D}_1(I)_i-\mathcal{D}_1(I)_j \right | )
\end{equation}
\vspace{-5mm}
\begin{equation}
    \mathcal{L}_{enh}= \omega_{exp}\frac{1}{K}\sum_{i=1}^{K}\left | I_{en,i}-E \right |+\omega_{col}\sum_{\forall(p,q)\in \varepsilon  }^{} (V_p-V_q)^2,\varepsilon =\{(R,G),(R,B),(G,B)\}
\end{equation}
The images before and after enhancement are divided into $K$ patches. Here, \(\sigma(i)\) represents the neighboring patches surrounding position \(i\). \(I_{en,i}\) and \(\mathcal{D}_1(I)_i\) denote the mean pixel values within the i-th patches at the corresponding position. The loss \( L_{enh} \) imposes constraints on the average brightness of the patches and the overall chromaticity of the image, where $V_p$ denotes the average intensity value of $p$ channel
in the enhanced image, and $E$ represents the exposure standard that aligns with natural perception. \( \omega_{exp} \) and \( \omega_{col} \) represent the respective weighting factors. Finally, the overall loss of the end-to-end network can be described as follows:
\begin{equation}
\mathcal{L}=\omega_R\mathcal{L}_{R}+\omega_L\mathcal{L}_{L}+\omega_{con}\mathcal{L}_{con}+\omega_{enh}\mathcal{L}_{enh}
\end{equation}
Here, \( \omega_R \), \( \omega_L \), \( \omega_{con} \), and \( \omega_{enh} \) represent the respective weighting factors.
\vspace{-2.5mm}
\begin{table}[h]\small
\centering
\caption{PSNR$\uparrow$, SSIM$\uparrow$, LPIPS$\downarrow$ scores on the image sets (LOLv1, LOLv2). The best result is in \textcolor{red}{red}, whereas the second-best one is in \textcolor{blue}{blue} under each case.}
\vspace{-2mm}
\label{tab:lol}
\begin{tabular}{cc|ccc|ccc}
\hline
\multicolumn{1}{l}{}  & \multicolumn{1}{l|}{} & \multicolumn{3}{c|}{LOLv1}                                                                                            & \multicolumn{3}{c}{LOLv2-Real}                                                                                        \\
Method                & Reference            & PSNR$\uparrow$                                  & SSIM$\uparrow$                                  & LPIPS$\downarrow$                                 & PSNR$\uparrow$                                  & SSIM$\uparrow$                                  & LPIPS$\downarrow$                                 \\ \hline
\textbf{Supervised}     & \multicolumn{1}{l|}{} & \multicolumn{1}{l}{}                  & \multicolumn{1}{l}{}                  & \multicolumn{1}{l|}{}                 & \multicolumn{1}{l}{}                  & \multicolumn{1}{l}{}                  & \multicolumn{1}{l}{}                  \\
URetinexNet          & \cite{wu2022uretinex}              & 19.84                                 & 0.824                                 & 0.237                                 & 21.09                                 & 0.858                                 & 0.208                                 \\
SNR-aware               & \cite{xu2022snr}                  & 24.61                                 & 0.842                                 & 0.233 & 21.48  & 0.849 & 0.237 \\
LLFormer                 & \cite{llmformer}                  & 23.65          & 0.818 & 0.169  & 27.75                                 & 0.861                                 & 0.142                                  \\ 
Retinexformer          & \cite{cai2023retinexformer}              & 23.93                                 & 0.831                                 & ——                                 & 21.23                                 & 0.838                                 & ——                                 \\
Retinexmamba          & \cite{retinexmamba}              & 24.03                                 & 0.831                                 & ——                                 & 22.45                                 & 0.844                                 & ——                                 \\
\hline
\textbf{Unpaired}     & \multicolumn{1}{l|}{} & \multicolumn{1}{l}{}                  & \multicolumn{1}{l}{}                  & \multicolumn{1}{l|}{}                 & \multicolumn{1}{l}{}                  & \multicolumn{1}{l}{}                  & \multicolumn{1}{l}{}                  \\
EnlightenGAN          & \cite{jiang2021enlightengan}              & 17.48                                 & 0.651                                 & 0.322                                 & 18.64                                 & 0.675                                 & 0.308                                 \\
PairLIE               & \cite{pairlie}                  & 19.51                                 & 0.736                                 & \textcolor{blue}{\textbf{0.247}} & \textcolor{blue}{\textbf{19.70}}  & \textcolor{blue}{\textbf{0.774}} & {\color[HTML]{FF0000} \textbf{0.235}} \\
Nerco                 & \cite{nerco}                  & {\color[HTML]{000000} 19.70}          & \textcolor{blue}{\textbf{0.742}} & {\color[HTML]{FF0000} \textbf{0.234}} & 19.66                                 & 0.717                                 & 0.270                                  \\ \hline
\textbf{No-Reference} & \multicolumn{1}{l|}{} & \multicolumn{1}{l}{}                  & \multicolumn{1}{l}{}                  & \multicolumn{1}{l|}{}                 & \multicolumn{1}{l}{}                  & \multicolumn{1}{l}{}                  & \multicolumn{1}{l}{}                  \\
ZERO-DCE              & \cite{ZERODCE}              & 14.86                                 & 0.559                                 & 0.335                                 & 18.06                                 & 0.573                                 & 0.312                                 \\
RUAS                  & \cite{RUAS}                   & 16.40                                  & 0.500                                   & 0.270                                  & 15.33                                 & 0.488                                 & 0.310                                  \\
Sci-easy              & \cite{SCI}                   & 9.58                                  & 0.369                                 & 0.410                                  & 11.98                                 & 0.399                                 & 0.354                                 \\
Sci-medium            &                   & 14.78                                 & 0.522                                 & 0.339                                 & 17.30                                  & 0.534                                 & 0.308                                 \\
Sci-hard              &               & 13.81                                 & 0.526                                 & 0.358                                 & 17.25                                 & 0.546                                 & 0.317                                 \\
Clip-LIT              & \cite{clip}              & 17.21                                 & 0.589                                 &0.335                                       & 17.06                                 & 0.589                                 &0.352                                       \\
Enlighten-Your-Voice  &\cite{eyv}               & \textcolor{blue}{\textbf{19.73}} & 0.715                                 &——                                       & 19.34                                 & 0.686                                 &——                                       \\
Ours                  &                   & {\color[HTML]{FF0000} \textbf{19.80}} & {\color[HTML]{FF0000} \textbf{0.750}} & 0.253                                 & {\color[HTML]{FF0000} \textbf{20.22}} & {\color[HTML]{FF0000} \textbf{0.793}} & \textcolor{blue}{\textbf{0.266}} \\ \hline
\end{tabular}
\vspace{1.5mm}
\centering
\caption{PSNR$\uparrow$/SSIM$\uparrow$/LPIPS$\downarrow$ scores on the image set SICE, and BRSIQUE$\downarrow$/CLIPIQA$\downarrow$ scores on the image set SIDD. The best result is in \textcolor{red}{red}, whereas the second-best one is in \textcolor{blue}{blue}.}
\vspace{-1mm}
\label{tab:s}
\begin{tabular}{cc|ccc|cc}
\hline
             &            & \multicolumn{3}{c|}{SICE}                                                                                             & \multicolumn{2}{c}{SIDD}                                                      \\
Method       & Parameters & PSNR$\uparrow$                                  & SSIM$\uparrow$                                  & LPIPS$\downarrow$                                 & BRSIQUE$\downarrow$                               & CLIPIQA$\downarrow$                               \\ \hline
\textbf{Supervised}     & \multicolumn{1}{l|}{} & \multicolumn{1}{l}{}                  & \multicolumn{1}{l}{}                  & \multicolumn{1}{l|}{}                 & \multicolumn{1}{l}{}                  & \multicolumn{1}{l}{}                  \\
URetinexNet          & 1.04M              & 22.12                                 & 0.844                                 & 0.462                                 & ——                                 & ——                                 \\
SNR-aware               & 50.95M                  & 15.02                                 & 0.584                                 & 0.527 & 25.679  & 0.294 \\
LLFormer                 & 72.29M                  & 17.88          & 0.821 & 0.503  & 3.548                                 & 0.339                                 \\
Retinexformer          & 1.61M              & ——                                 & ——                                 & ——                                 & 9.229                                & 0.343                                \\
Retinexmamba          &4.59M              & ——                                 & ——                                 & ——                                 & 11.826                                 & 0.386                                \\
\hline
\textbf{Unpaired}     &            &                                       &                                       &       &                                       &                                       \\
EnlightenGAN & 8.44M   & 18.73          & 0.822          & {\color[HTML]{FE0000} \textbf{0.216}} & 13.786                                & \textcolor{blue}{\textbf{0.337}} \\
PairLIE      & 0.34M       & \textcolor{blue}{\textbf{21.32}} & \textcolor{blue}{ \textbf{0.840}}  & {\color[HTML]{FE0000} \textbf{0.216}} & \textcolor{blue}{ \textbf{3.168}} & 0.383                                 \\
Nerco        &22.76M       & 18.72&0.805&0.474          & ——                                    & ——            \\ \hline
\textbf{No-Reference} &            &                &                &               &                                       &                                       \\
ZERO-DCE     & 0.08M   & 18.69&0.810&0.279          & 24.291                                & 0.503                                 \\
RUAS         & 0.01M        & 13.18& 0.734& 0.363          & 31.613                                & 0.361                                 \\
Sci-easy     & 0.01M        & 11.71&0.590&0.502          & 25.344                                & 0.399                                 \\
Sci-medium   &        &  15.95& 0.787& 0.335          & 21.636                                & 0.456                                 \\
Sci-hard     &    &  17.59& 0.782          &  0.486          & 35.533                                & 0.508                                 \\
Clip-LIT     & 0.27M   &  13.70          &  0.725          &  0.480          & 31.093                                & 0.434                                 \\
Ours         & 0.36M       & {\color[HTML]{FE0000} \textbf{22.55}} & {\color[HTML]{FE0000} \textbf{0.841}} & \textcolor{blue}{\textbf{0.234}} & {\color[HTML]{FE0000} \textbf{2.555}} & {\color[HTML]{FE0000} \textbf{0.292}} \\ \hline
\end{tabular}
\vspace{-5mm}
\end{table}
\begin{figure}[h]
    \centering
    \begin{minipage}{\textwidth}
        \centering
        \includegraphics[width=\textwidth]{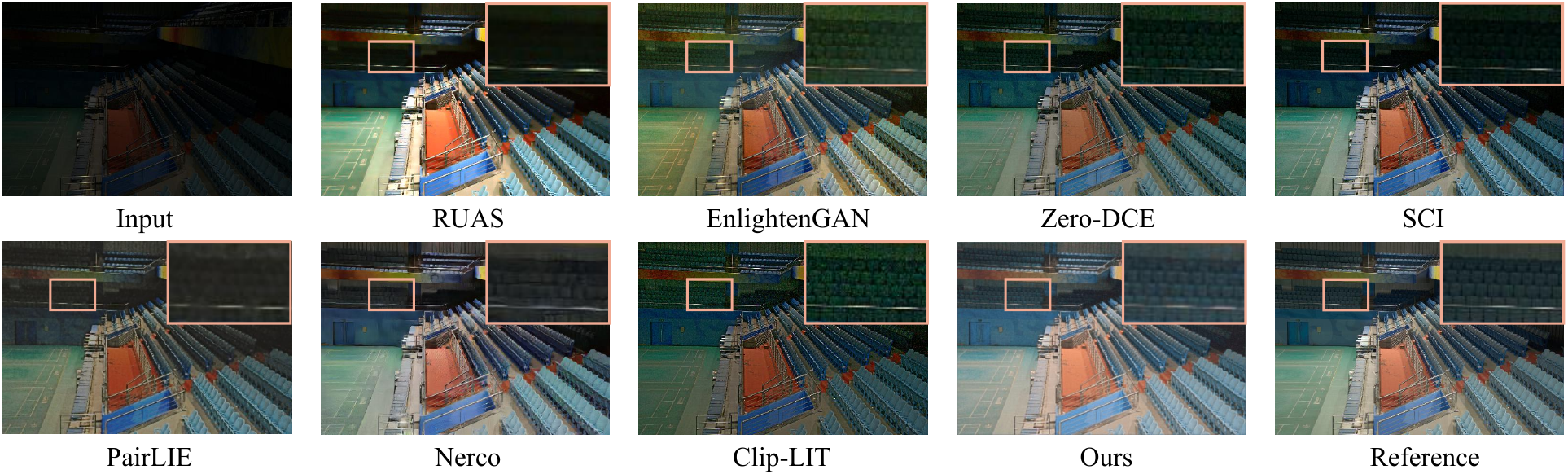}
        \caption{Visual comparison of typical unsupervised enhancement methods in LOL~\cite{lolv2}. Flesh pink boxes indicate the obvious differences.}
        \label{fig:lol}
    \end{minipage}
    %\vspace{0.5cm} % Optional vertical space between images
    \begin{minipage}{\textwidth}
        \centering
        \includegraphics[width=\textwidth]{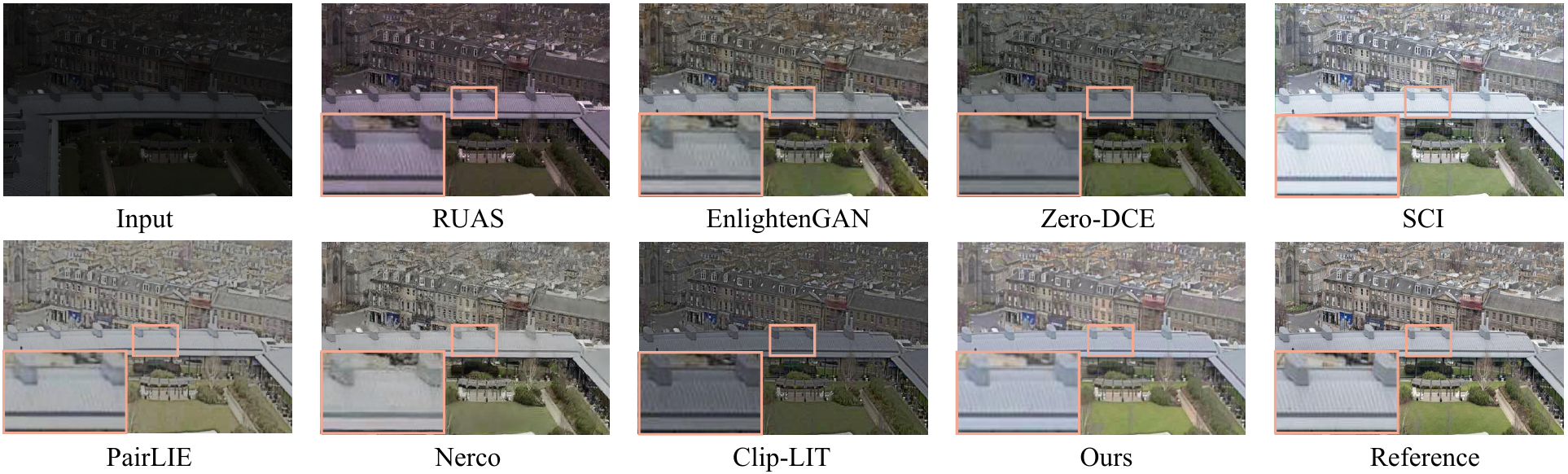}
        \caption{Visual comparison of typical unsupervised enhancement methods in SICE~\cite{sice}. Flesh pink boxes indicate the obvious differences.}
        \label{fig:sice}
    \end{minipage}
    \begin{minipage}{\textwidth}
        \centering
        \includegraphics[width=\textwidth]{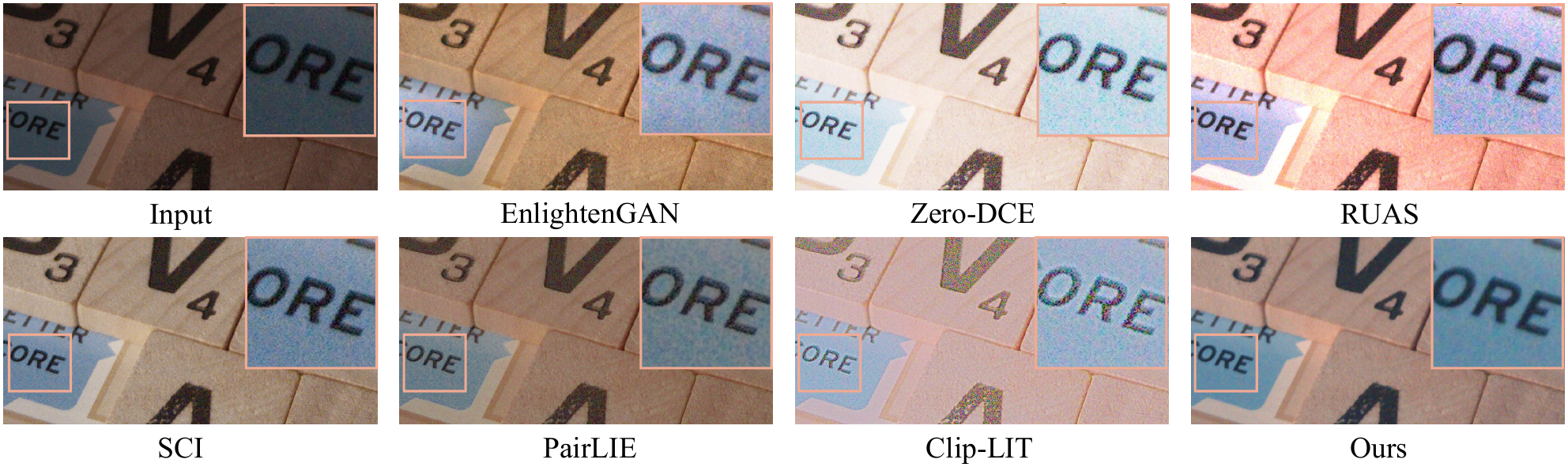}
        \caption{Visual comparison on the real-world low-light image from the SIDD \cite{sidd} dataset.}
    \vspace{-6mm}
    \label{fig:sidd}
    \end{minipage}
\end{figure}
\vspace{-2mm}
\section{experiment}
\vspace{-1.5mm}
\subsection{implementation details}
\vspace{-2.5mm}
To ensure fairness, all experiments were terminated after 100 training epochs. We consistently set the initial learning rate to \( 1 \times 10^{-5} \) and conducted all experiments on an RTX 3090 GPU. During training, images were randomly cropped into 256x256 patches, with pixel values normalized to the range of (0, 1), and a batch size of 1 was employed.

We conducted tests on four benchmarks: LOLv1~\cite{lolv1}, LOLv2-real~\cite{lolv2}, SICE~\cite{sice} and SIDD~\cite{sidd}. Please refer to the supplementary materials for detailed information regarding the datasets, including the corresponding training and testing splits.
\vspace{-4mm}
\subsection{benchmarking results}
\vspace{-4mm}
The experimental results on the LOL dataset are presented in Tab.~\ref{tab:lol}, where our model outperforms most of the compared unpaired and no-reference methods, achieving the highest scores across multiple metrics. The qualitative visual comparisons are shown in Fig.~\ref{fig:lol}. Unpaired methods benefit from reference images captured in normal lighting conditions, making learning the necessary illumination features easier. However, these methods struggle with underexposed local regions, leading to issues like dead black spots.

Meanwhile, EnlightGAN, ZeroDCE, and Clip-LIT successfully enhance overly dark regions. However, due to the lack of proper denoising mechanisms, they tend to introduce noise while increasing exposure. Our approach, leveraging illumination priors and frequency domain decomposition, effectively compensates for multidimensional illumination information, resolving complex degradation issues such as local overexposure, underexposure, and noise.

The experimental results on the SICE and SIDD datasets are shown in Tab.~\ref{tab:s}. The selected SICE test set includes images with three levels of low-light degradation: low, medium, and high. We evaluate the generalization capability of our model under varying illumination conditions using statistical metrics, and the qualitative comparisons are shown in Fig.~\ref{fig:sice}. Both RUAS and EnlightenGAN exhibit issues such as local overexposure and strong contrast distortion, which can be attributed to the lack of an interpretable illumination feedback design in their network structures. Nerco generates artifacts in certain image regions, highlighting the uncontrollability of generative models in image enhancement tasks. In contrast, our method demonstrates appropriate contrast, accurate chrominance, low noise, and sufficient detail.

The qualitative comparison results on the SIDD dataset are shown in Fig.~\ref{fig:sidd}. We assess the enhancement capability of our model in challenging low-light scenes with high noise levels and complex noise patterns. Our method achieves the best performance on two no-reference statistical metrics, BRISQUE and CLIPIQA, indicating that the enhanced images exhibit characteristics closer to natural images with fewer distortions. From the visual results, our method demonstrates robustness against complex noise in real-world scenarios, effectively enhancing image illumination while controlling noise intensity. In contrast, other approaches either lack a dedicated denoising design or handle noise from a perceptual standpoint, without corresponding theoretical analysis for interpretability, leading to suboptimal results.
\begin{table}[h]\small
\centering
\caption{Ablation study of the contribution of the three physical priors. The best and the second best results are highlighted in \textcolor{red}{red} and \textcolor{blue}{blue}.}
\vspace{-2mm}
\label{tab:prior}
\begin{tabular}{ccc|ccc|ccc}
\hline
        &    &      & \multicolumn{3}{c|}{LOLv1} & \multicolumn{3}{c}{LOLv2} \\
Illumination&Lowpass & Highpass & PSNR$\uparrow$    & SSIM$\uparrow$    & LPIPS$\downarrow$  & PSNR$\uparrow$    & SSIM$\uparrow$   & LPIPS$\downarrow$  \\ \hline
        $\times$         &$\times$& $\times$         & 18.88   & 0.741   & 0.273  & 19.37   &  0.771  & 0.305  \\
        $\checkmark$         &$\times$& $\times$         & 19.54   & {\color[HTML]{FE0000} \textbf{0.753}}   & {\color[HTML]{FE0000} \textbf{0.253}}  & \textcolor{blue}{ \textbf{19.99}}   & \textcolor{blue}{\textbf{0.785}}  & 0.297  \\
      $\checkmark$         &$\times$  & $\checkmark$         & \textcolor{blue} {\textbf{19.69}}   & 0.744   & \textcolor{blue}{\textbf{0.259}}  & 19.28   & 0.779  & 0.299  \\
        $\checkmark$         &$\checkmark$ &  $\times$        & 19.57   & 0.745   & 0.262  & 19.51   & 0.780   & \textcolor{blue}{\textbf{0.282}}  \\
        $\checkmark$         &$\checkmark$ & $\checkmark$          & {\color[HTML]{FE0000} \textbf{19.80}}    & \textcolor{blue}{\textbf{0.750}}    & {\color[HTML]{FE0000} \textbf{0.253}}  & {\color[HTML]{FE0000} \textbf{20.22}}   & {\color[HTML]{FE0000} \textbf{0.793}}  & {\color[HTML]{FE0000} \textbf{0.266}}  \\ \hline
\end{tabular}
\vspace{3mm}
\centering
\caption{Ablation study of the contribution of the denoising designs, where NM stands for neighborhood masking. The best and the second best results are highlighted in \textcolor{red}{red} and \textcolor{blue}{blue}.}
\vspace{-2mm}
\label{tab:den}
\begin{tabular}{ccc|ccc|ccc}
\hline
                & &      & \multicolumn{3}{c|}{LOLv1} & \multicolumn{3}{c}{LOLv2} \\
Setting&NM & $\mathcal{L}_{reg}$ & PSNR$\uparrow$    & SSIM$\uparrow$    & LPIPS$\downarrow$  & PSNR$\uparrow$    & SSIM$\uparrow$   & LPIPS$\downarrow$  \\ \hline
                 1&$\times$ & $\times$      & 18.52   & 0.686   & 0.271  & 19.46   & 0.771  & 0.323  \\
                 2&$\checkmark$ & $\times$     & 19.63   & 0.747   & 0.264  & 19.83   & 0.787  & 0.279  \\
                 3&$\checkmark$ & $\checkmark$      & {\color[HTML]{FE0000} \textbf{19.80}}   & {\color[HTML]{FE0000} \textbf{0.750}}   & {\color[HTML]{FE0000} \textbf{0.253}}  & {\color[HTML]{FE0000} \textbf{20.22}}   & {\color[HTML]{FE0000} \textbf{0.793}}  & {\color[HTML]{FE0000} \textbf{0.266}}  \\ \hline
\end{tabular}
\vspace{-3mm}
\end{table}
\vspace{-2mm}
\subsection{ablation study}
\vspace{-2mm}
\textbf{Denoiseing Design.} In the previous sections, we designed a hybrid mechanism combining neighborhood masking and gamma enhancement to construct image pairs with varying illumination and noise levels for joint denoising and enhancement training. In set1, we removed the masking mechanism and trained using the original resolution images with different illumination. In set2, we applied the full preprocessing mechanism but omitted the regularization term in Equ.~\ref{equ:reg}. 

\begin{figure}[h]
    \centering
    \begin{subfigure}{0.44\textwidth}
        \centering
        \includegraphics[width=\linewidth]{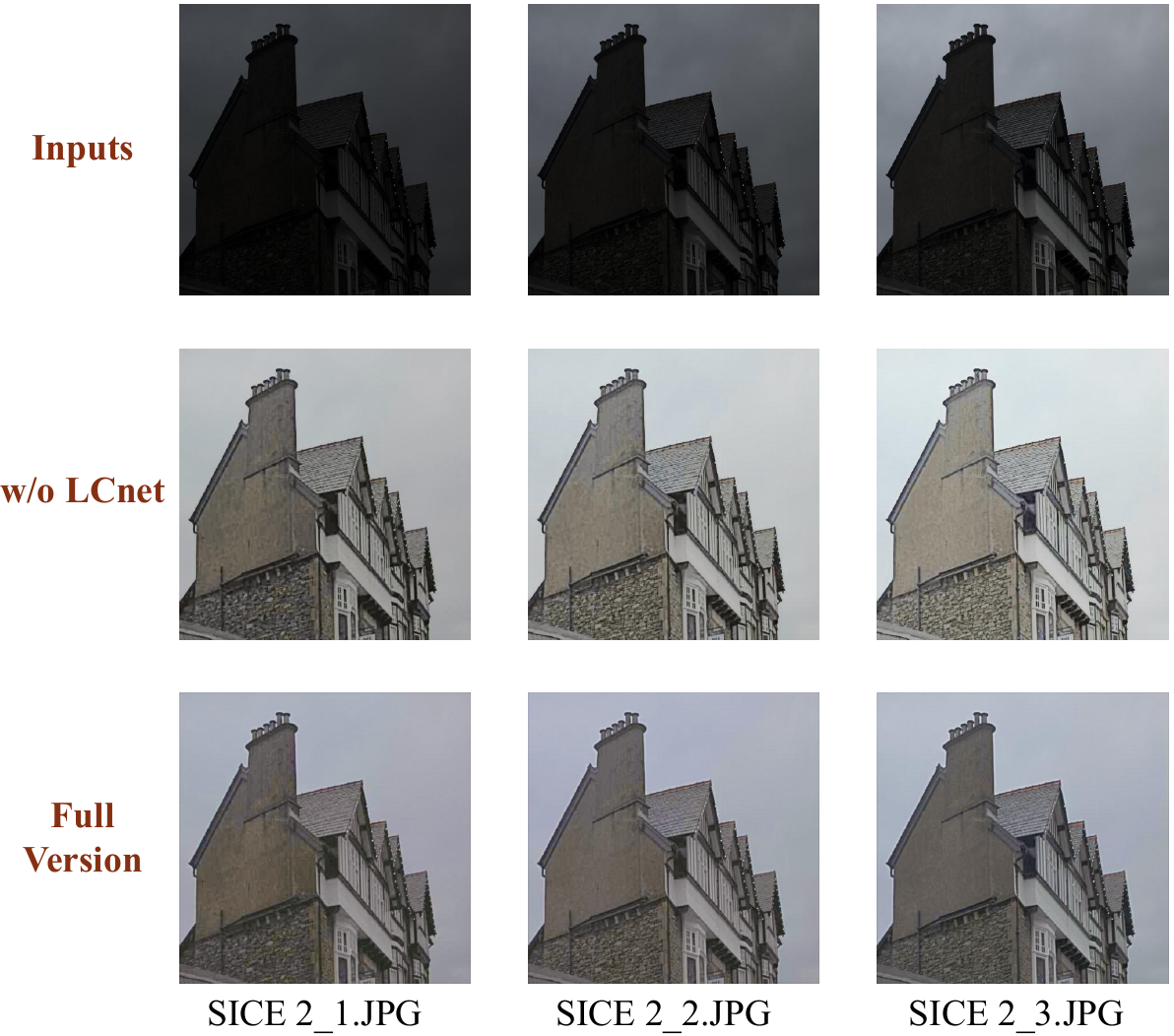}
        \label{fig:first}
    \end{subfigure}
    \hfill
    \begin{subfigure}{0.52\textwidth}
        \centering
        \includegraphics[width=\linewidth]{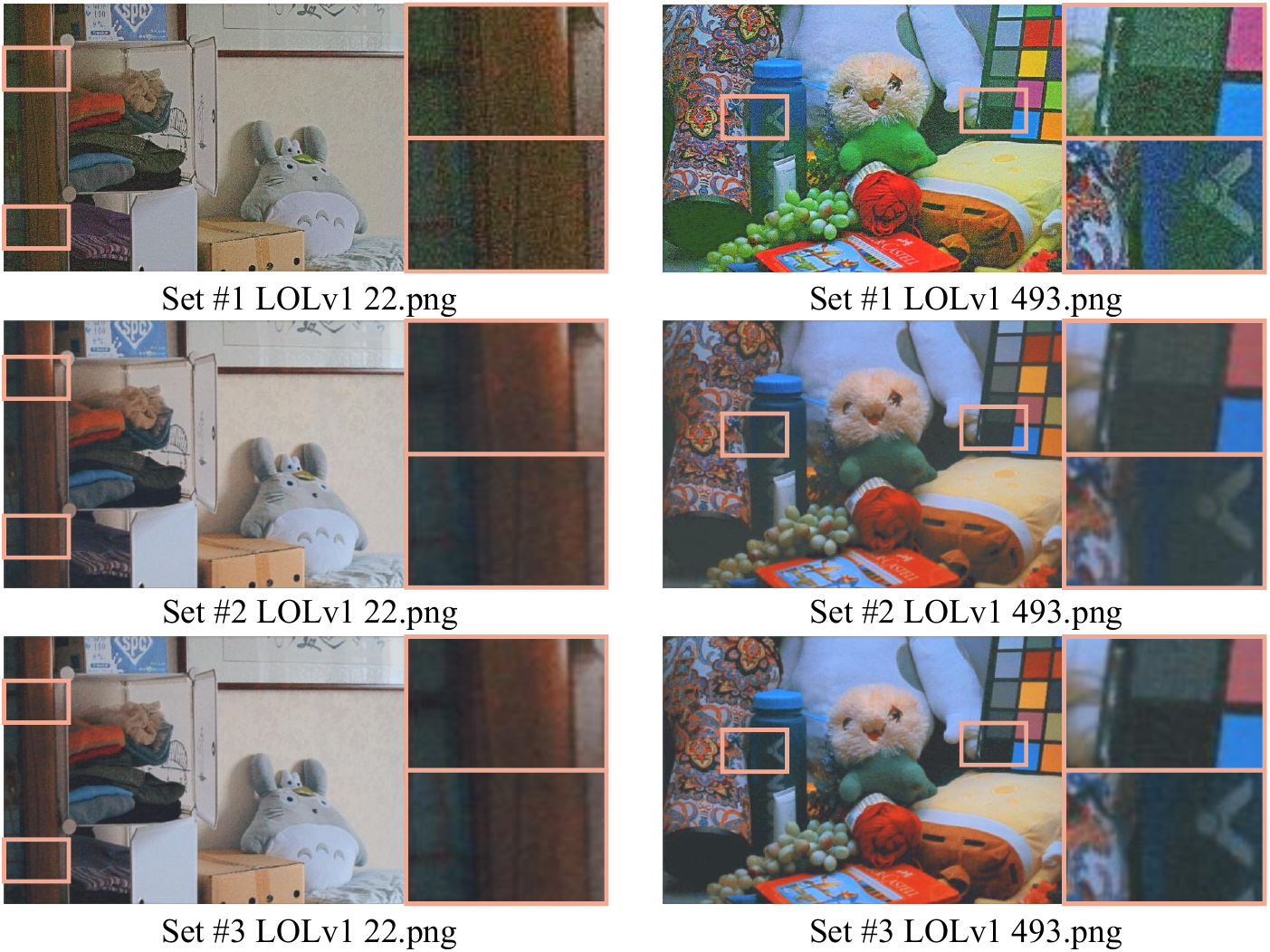}
        \label{fig:second}
    \end{subfigure}
    \vspace{-4mm}
    \caption{Left: Visualization of LCnet adaptivity experiment. Right: Visualization of denoising design ablation.}
    \vspace{-3mm}
    \label{fig:overall1}
\end{figure}

We implemented these settings on the LOLv1 and LOLv2-Real datasets, with the quantitative results presented in Tab.~\ref{tab:den} and the visual comparisons shown in Fig.\ref{fig:overall1}.

The results indicate that removing any part of the strategy reduces performance, and the combination of both strategies is necessary to achieve optimal denoising results. In set1, the noise intensity is significantly pronounced, primarily due to the decomposition network generating identity mappings while learning the illumination map. In set2, images lose detail in underexposed regions, which is attributed to the local semantic loss caused by downsampling.
\begin{figure}[h]
    \centering
    \begin{subfigure}{0.25\textwidth}
        \centering
        \includegraphics[width=\linewidth]{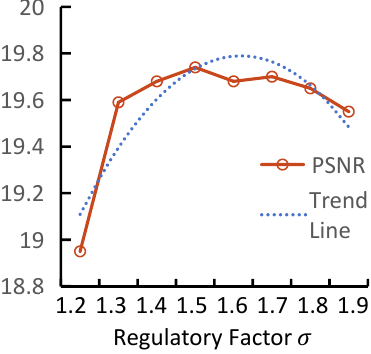}
        \label{fig:first}
    \end{subfigure}
    \hfill
    \begin{subfigure}{0.72\textwidth}
        \centering
        \includegraphics[width=\linewidth]{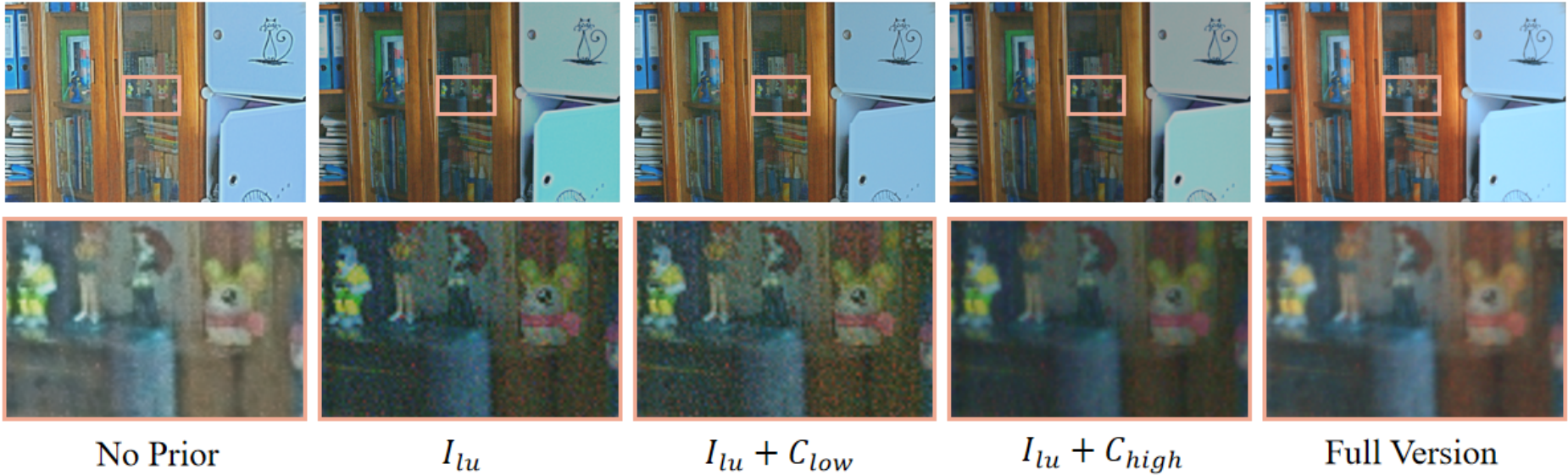}
        \label{fig:second}
    \end{subfigure}
    \vspace{-4mm}
    \caption{Left: PSNR variation with gamma enhancement factor on the LOLv1 dataset. Right: Ablation study of different physical priors.}
    \vspace{-2mm}
    \label{fig:overall2}
    \vspace{-2.5mm}
\end{figure}

\textbf{Hybrid Piror Design.} Tab.~\ref{tab:prior} and Fig.~\ref{fig:overall2} present the results of the ablation study on mixed priors. The priors are categorized into three parts: illumination prior, high-pass filtering prior, and low-pass filtering prior, which respectively capture brightness, noise, and color information. The results are worse when all priors are removed, with a notable improvement of approximately 0.6 dB when the illumination prior is included. On top of the full version, removing either the high-frequency or low-frequency components adversely affects performance, demonstrating that combining multiple informative cues achieves the best results.

\textbf{LCnet.} The design of LCnet aims to build an illumination-adaptive module that adjusts the illumination map to achieve the highest perceptual quality. We removed LCnet and employed a reference adjustment strategy similar to PairLIE. The visual results are shown in Fig.~\ref{fig:overall1}. Without the adaptive strategy, it is challenging to achieve consistent enhancement results across images with varying low-light degradations from the same scene, leading to overexposure in local regions.

\textbf{Gamma Enhancement Factor.} For the gamma enhancement operation applied to images with different illumination during pre-training, we explored which enhancement factor yields the best performance. We use $\sigma$ to regulate $\lambda$ through the formula $\lambda = \frac{1}{\sigma}$. The results are shown in Fig.~\ref{fig:overall2}, illustrating that the enhancement effect follows an increasing trend initially and then decreases within the $\sigma$ range of 1.2 to 1.9. At lower values, the enhanced images do not produce sufficient illumination differences with the other sub-images, which is crucial for model decomposition. At higher values, the enhancement does not conform to the assumption \( R_1^{\lambda-1}=1 \) during framework inference, resulting in more complex nonlinear noise variations that negatively impact model performance. Therefore, during each iteration, we randomly sample enhancement factors within the range of $(1.3, 1.7)$ to provide the model with a broader range of feature processing options. The specific selection criteria for the control factors are detailed in the supplementary materials.
\vspace{-2.5mm}
\section{conclusion}
This paper tackles the challenges of low-light image enhancement and denoising, particularly in complex real-world scenarios. We propose a zero-reference framework combining self-supervised denoising via neighboring pixel downsampling and enhancement using random gamma adjustment with retinal perception theory. To address the limitations of existing methods in handling frequency-domain degradations, we introduce an RGB-space DCT-based filtering module for multi-frequency separation and a Dynamic Discrete Sequence Fusion Transformer to integrate frequency-domain priors. Experiments on real-world datasets show our method outperforms state-of-the-art techniques, offering a robust solution for low-light enhancement and denoising.

\bibliography{iclr2025_conference}

\begin{thebibliography}{42}
\providecommand{\natexlab}[1]{#1}
\providecommand{\url}[1]{\texttt{#1}}
\expandafter\ifx\csname urlstyle\endcsname\relax
  \providecommand{\doi}[1]{doi: #1}\else
  \providecommand{\doi}{doi: \begingroup \urlstyle{rm}\Url}\fi

\bibitem[Abdelhamed et~al.(2018)Abdelhamed, Lin, and Brown]{sidd}
Abdelrahman Abdelhamed, Stephen Lin, and Michael~S Brown.
\newblock A high-quality denoising dataset for smartphone cameras.
\newblock In \emph{Proceedings of the IEEE conference on computer vision and pattern recognition}, pp.\  1692--1700, 2018.

\bibitem[Bai et~al.(2024)Bai, Yin, He, Li, and Zhang]{retinexmamba}
Jiesong Bai, Yuhao Yin, Qiyuan He, Yuanxian Li, and Xiaofeng Zhang.
\newblock Retinexmamba: Retinex-based mamba for low-light image enhancement.
\newblock \emph{arXiv preprint arXiv:2405.03349}, 2024.

\bibitem[Cai et~al.(2018)Cai, Gu, and Zhang]{sice}
Jianrui Cai, Shuhang Gu, and Lei Zhang.
\newblock Learning a deep single image contrast enhancer from multi-exposure images.
\newblock \emph{IEEE Transactions on Image Processing}, 27\penalty0 (4):\penalty0 2049--2062, 2018.

\bibitem[Cai et~al.(2021)Cai, Zhang, Huang, Geng, Li, and Huang]{cai2021frequency}
Mu~Cai, Hong Zhang, Huijuan Huang, Qichuan Geng, Yixuan Li, and Gao Huang.
\newblock Frequency domain image translation: More photo-realistic, better identity-preserving.
\newblock In \emph{Proceedings of the IEEE/CVF International Conference on Computer Vision}, pp.\  13930--13940, 2021.

\bibitem[Cai et~al.(2023)Cai, Bian, Lin, Wang, Timofte, and Zhang]{cai2023retinexformer}
Yuanhao Cai, Hao Bian, Jing Lin, Haoqian Wang, Radu Timofte, and Yulun Zhang.
\newblock Retinexformer: One-stage retinex-based transformer for low-light image enhancement.
\newblock In \emph{Proceedings of the IEEE/CVF International Conference on Computer Vision}, pp.\  12504--12513, 2023.

\bibitem[Ch{\k{e}}i{\'n}ski \& Wawrzy{\'n}ski(2020)Ch{\k{e}}i{\'n}ski and Wawrzy{\'n}ski]{dctconv}
Karol Ch{\k{e}}i{\'n}ski and Pawe{\l} Wawrzy{\'n}ski.
\newblock Dct-conv: Coding filters in convolutional networks with discrete cosine transform.
\newblock In \emph{2020 International Joint Conference on Neural Networks (IJCNN)}, pp.\  1--6. IEEE, 2020.

\bibitem[Chen et~al.(2021)Chen, Yan, Deng, Wu, Wu, Xu, and Zhou]{tianwen}
Xuelei Chen, Jingye Yan, Li~Deng, Fengquan Wu, Lin Wu, Yidong Xu, and Li~Zhou.
\newblock Discovering the sky at the longest wavelengths with a lunar orbit array.
\newblock \emph{Philosophical Transactions of the Royal Society A}, 379\penalty0 (2188):\penalty0 20190566, 2021.

\bibitem[Fan et~al.(2022)Fan, Fan, Gan, Chen, and Chen]{fan2022multiscale}
Guo-Dong Fan, Bi~Fan, Min Gan, Guang-Yong Chen, and CL~Philip Chen.
\newblock Multiscale low-light image enhancement network with illumination constraint.
\newblock \emph{IEEE Transactions on Circuits and Systems for Video Technology}, 32\penalty0 (11):\penalty0 7403--7417, 2022.

\bibitem[Fu et~al.(2023)Fu, Yang, Tu, Huang, Ding, and Ma]{pairlie}
Zhenqi Fu, Yan Yang, Xiaotong Tu, Yue Huang, Xinghao Ding, and Kai-Kuang Ma.
\newblock Learning a simple low-light image enhancer from paired low-light instances.
\newblock In \emph{Proceedings of the IEEE/CVF conference on computer vision and pattern recognition}, pp.\  22252--22261, 2023.

\bibitem[Gao et~al.(2024)Gao, Xu, Zhao, and Liu]{fcdiffusion}
Xiang Gao, Zhengbo Xu, Junhan Zhao, and Jiaying Liu.
\newblock Frequency-controlled diffusion model for versatile text-guided image-to-image translation.
\newblock In \emph{Proceedings of the AAAI Conference on Artificial Intelligence}, volume~38, pp.\  1824--1832, 2024.

\bibitem[Goyal et~al.(2020)Goyal, Dogra, Agrawal, Sohi, and Sharma]{denoisingreview}
Bhawna Goyal, Ayush Dogra, Sunil Agrawal, Balwinder~Singh Sohi, and Apoorav Sharma.
\newblock Image denoising review: From classical to state-of-the-art approaches.
\newblock \emph{Information fusion}, 55:\penalty0 220--244, 2020.

\bibitem[Guo et~al.(2020)Guo, Li, Guo, Loy, Hou, Kwong, and Cong]{ZERODCE}
Chunle Guo, Chongyi Li, Jichang Guo, Chen~Change Loy, Junhui Hou, Sam Kwong, and Runmin Cong.
\newblock Zero-reference deep curve estimation for low-light image enhancement.
\newblock In \emph{Proceedings of the IEEE/CVF conference on computer vision and pattern recognition}, pp.\  1780--1789, 2020.

\bibitem[Huang et~al.(2012)Huang, Cheng, and Chiu]{gamma}
Shih-Chia Huang, Fan-Chieh Cheng, and Yi-Sheng Chiu.
\newblock Efficient contrast enhancement using adaptive gamma correction with weighting distribution.
\newblock \emph{IEEE transactions on image processing}, 22\penalty0 (3):\penalty0 1032--1041, 2012.

\bibitem[Jiang et~al.(2021)Jiang, Gong, Liu, Cheng, Fang, Shen, Yang, Zhou, and Wang]{jiang2021enlightengan}
Yifan Jiang, Xinyu Gong, Ding Liu, Yu~Cheng, Chen Fang, Xiaohui Shen, Jianchao Yang, Pan Zhou, and Zhangyang Wang.
\newblock Enlightengan: Deep light enhancement without paired supervision.
\newblock \emph{IEEE transactions on image processing}, 30:\penalty0 2340--2349, 2021.

\bibitem[Jin et~al.(2022)Jin, Yang, and Tan]{nightenhance}
Yeying Jin, Wenhan Yang, and Robby~T Tan.
\newblock Unsupervised night image enhancement: When layer decomposition meets light-effects suppression.
\newblock In \emph{European Conference on Computer Vision}, pp.\  404--421. Springer, 2022.

\bibitem[Jin et~al.(2023)Jin, Lin, Yan, Yuan, Ye, and Tan]{apsf}
Yeying Jin, Beibei Lin, Wending Yan, Yuan Yuan, Wei Ye, and Robby~T Tan.
\newblock Enhancing visibility in nighttime haze images using guided apsf and gradient adaptive convolution.
\newblock In \emph{Proceedings of the 31st ACM international conference on multimedia}, pp.\  2446--2457, 2023.

\bibitem[Land \& McCann(1971)Land and McCann]{retinex}
Edwin~H Land and John~J McCann.
\newblock Lightness and retinex theory.
\newblock \emph{Josa}, 61\penalty0 (1):\penalty0 1--11, 1971.

\bibitem[Lee et~al.(2013)Lee, Lee, and Kim]{histogram}
Chulwoo Lee, Chul Lee, and Chang-Su Kim.
\newblock Contrast enhancement based on layered difference representation of 2d histograms.
\newblock \emph{IEEE transactions on image processing}, 22\penalty0 (12):\penalty0 5372--5384, 2013.

\bibitem[Lehtinen et~al.(2018)Lehtinen, Munkberg, Hasselgren, Laine, Karras, Aittala, and Aila]{n2n}
Jaakko Lehtinen, Jacob Munkberg, Jon Hasselgren, Samuli Laine, Tero Karras, Miika Aittala, and Timo Aila.
\newblock Noise2noise: Learning image restoration without clean data (2018).
\newblock \emph{arXiv preprint arXiv:1803.04189}, 2018.

\bibitem[Li et~al.(2021)Li, Guo, and Loy]{zerodceplus}
Chongyi Li, Chunle Guo, and Chen~Change Loy.
\newblock Learning to enhance low-light image via zero-reference deep curve estimation.
\newblock \emph{IEEE transactions on pattern analysis and machine intelligence}, 44\penalty0 (8):\penalty0 4225--4238, 2021.

\bibitem[Li et~al.(2024)Li, Li, Tu, Liu, Guo, Juefei-Xu, Xu, and Yu]{lightdriving}
Jinlong Li, Baolu Li, Zhengzhong Tu, Xinyu Liu, Qing Guo, Felix Juefei-Xu, Runsheng Xu, and Hongkai Yu.
\newblock Light the night: A multi-condition diffusion framework for unpaired low-light enhancement in autonomous driving.
\newblock In \emph{Proceedings of the IEEE/CVF Conference on Computer Vision and Pattern Recognition}, pp.\  15205--15215, 2024.

\bibitem[Liang et~al.(2023)Liang, Li, Zhou, Feng, and Loy]{clip}
Zhexin Liang, Chongyi Li, Shangchen Zhou, Ruicheng Feng, and Chen~Change Loy.
\newblock Iterative prompt learning for unsupervised backlit image enhancement.
\newblock In \emph{Proceedings of the IEEE/CVF International Conference on Computer Vision}, pp.\  8094--8103, 2023.

\bibitem[Liu et~al.(2023)Liu, Wang, Li, Wang, and Qian]{liu2023fsi}
Chengxu Liu, Xuan Wang, Shuai Li, Yuzhi Wang, and Xueming Qian.
\newblock Fsi: Frequency and spatial interactive learning for image restoration in under-display cameras.
\newblock In \emph{Proceedings of the IEEE/CVF International Conference on Computer Vision}, pp.\  12537--12546, 2023.

\bibitem[Liu et~al.(2021)Liu, Ma, Zhang, Fan, and Luo]{RUAS}
Risheng Liu, Long Ma, Jiaao Zhang, Xin Fan, and Zhongxuan Luo.
\newblock Retinex-inspired unrolling with cooperative prior architecture search for low-light image enhancement.
\newblock In \emph{Proceedings of the IEEE/CVF conference on computer vision and pattern recognition}, pp.\  10561--10570, 2021.

\bibitem[Ma et~al.(2022)Ma, Ma, Liu, Fan, and Luo]{SCI}
Long Ma, Tengyu Ma, Risheng Liu, Xin Fan, and Zhongxuan Luo.
\newblock Toward fast, flexible, and robust low-light image enhancement.
\newblock In \emph{Proceedings of the IEEE/CVF conference on computer vision and pattern recognition}, pp.\  5637--5646, 2022.

\bibitem[Ou et~al.(2024)Ou, Zhao, Guo, Zhang, and Lin]{ou2024winnet}
Wenjie Ou, Zhishuo Zhao, Dongyue Guo, Zheng Zhang, and Yi~Lin.
\newblock Winnet: Make only one convolutional layer effective for time series forecasting.
\newblock In \emph{International Conference on Intelligent Computing}, pp.\  348--359. Springer, 2024.

\bibitem[Rashed et~al.(2019)Rashed, Ramzy, Vaquero, El~Sallab, Sistu, and Yogamani]{objectdetection}
Hazem Rashed, Mohamed Ramzy, Victor Vaquero, Ahmad El~Sallab, Ganesh Sistu, and Senthil Yogamani.
\newblock Fusemodnet: Real-time camera and lidar based moving object detection for robust low-light autonomous driving.
\newblock In \emph{Proceedings of the IEEE/CVF International Conference on Computer Vision Workshops}, pp.\  0--0, 2019.

\bibitem[Serengil \& Ozpinar(2020)Serengil and Ozpinar]{lightface}
Sefik~Ilkin Serengil and Alper Ozpinar.
\newblock Lightface: A hybrid deep face recognition framework.
\newblock In \emph{2020 innovations in intelligent systems and applications conference (ASYU)}, pp.\  1--5. IEEE, 2020.

\bibitem[Wang et~al.(2022)Wang, Chen, Cai, Chen, Li, Sotelo, and Li]{segmetation}
Hai Wang, Yanyan Chen, Yingfeng Cai, Long Chen, Yicheng Li, Miguel~Angel Sotelo, and Zhixiong Li.
\newblock Sfnet-n: An improved sfnet algorithm for semantic segmentation of low-light autonomous driving road scenes.
\newblock \emph{IEEE Transactions on Intelligent Transportation Systems}, 23\penalty0 (11):\penalty0 21405--21417, 2022.

\bibitem[Wang et~al.(2023)Wang, Zhang, Shen, Luo, Stenger, and Lu]{llmformer}
Tao Wang, Kaihao Zhang, Tianrun Shen, Wenhan Luo, Bjorn Stenger, and Tong Lu.
\newblock Ultra-high-definition low-light image enhancement: A benchmark and transformer-based method.
\newblock In \emph{Proceedings of the AAAI Conference on Artificial Intelligence}, volume~37, pp.\  2654--2662, 2023.

\bibitem[Wei et~al.(2018)Wei, Wang, Yang, and Liu]{lolv1}
Chen Wei, Wenjing Wang, Wenhan Yang, and Jiaying Liu.
\newblock Deep retinex decomposition for low-light enhancement.
\newblock \emph{arXiv preprint arXiv:1808.04560}, 2018.

\bibitem[Wu et~al.(2022)Wu, Weng, Zhang, Wang, Yang, and Jiang]{wu2022uretinex}
Wenhui Wu, Jian Weng, Pingping Zhang, Xu~Wang, Wenhan Yang, and Jianmin Jiang.
\newblock Uretinex-net: Retinex-based deep unfolding network for low-light image enhancement.
\newblock In \emph{Proceedings of the IEEE/CVF conference on computer vision and pattern recognition}, pp.\  5901--5910, 2022.

\bibitem[Xie et~al.(2021)Xie, Song, Xu, Xu, Zhang, and Wang]{xie2021learning}
Wenbin Xie, Dehua Song, Chang Xu, Chunjing Xu, Hui Zhang, and Yunhe Wang.
\newblock Learning frequency-aware dynamic network for efficient super-resolution.
\newblock In \emph{Proceedings of the IEEE/CVF International Conference on Computer Vision}, pp.\  4308--4317, 2021.

\bibitem[Xing et~al.(2024)Xing, Hu, Metzen, Groh, Karaoglu, and Gevers]{retinex-diffusion}
Xiaoyan Xing, Vincent~Tao Hu, Jan~Hendrik Metzen, Konrad Groh, Sezer Karaoglu, and Theo Gevers.
\newblock Retinex-diffusion: On controlling illumination conditions in diffusion models via retinex theory.
\newblock \emph{arXiv preprint arXiv:2407.20785}, 2024.

\bibitem[Xu et~al.(2022)Xu, Wang, Fu, and Jia]{xu2022snr}
Xiaogang Xu, Ruixing Wang, Chi-Wing Fu, and Jiaya Jia.
\newblock Snr-aware low-light image enhancement.
\newblock In \emph{Proceedings of the IEEE/CVF conference on computer vision and pattern recognition}, pp.\  17714--17724, 2022.

\bibitem[Yang et~al.(2023)Yang, Ding, Wu, Li, and Zhang]{nerco}
Shuzhou Yang, Moxuan Ding, Yanmin Wu, Zihan Li, and Jian Zhang.
\newblock Implicit neural representation for cooperative low-light image enhancement.
\newblock In \emph{Proceedings of the IEEE/CVF international conference on computer vision}, pp.\  12918--12927, 2023.

\bibitem[Yang et~al.(2021)Yang, Wang, Huang, Wang, and Liu]{lolv2}
Wenhan Yang, Wenjing Wang, Haofeng Huang, Shiqi Wang, and Jiaying Liu.
\newblock Sparse gradient regularized deep retinex network for robust low-light image enhancement.
\newblock \emph{IEEE Transactions on Image Processing}, 30:\penalty0 2072--2086, 2021.

\bibitem[Zhang et~al.(2017)Zhang, Zuo, Chen, Meng, and Zhang]{dncnn}
Kai Zhang, Wangmeng Zuo, Yunjin Chen, Deyu Meng, and Lei Zhang.
\newblock Beyond a gaussian denoiser: Residual learning of deep cnn for image denoising.
\newblock \emph{IEEE transactions on image processing}, 26\penalty0 (7):\penalty0 3142--3155, 2017.

\bibitem[Zhang et~al.(2023)Zhang, Xu, Tang, Gu, Chen, Zhu, and Guan]{eyv}
Xiaofeng Zhang, Zishan Xu, Hao Tang, Chaochen Gu, Wei Chen, Shanying Zhu, and Xinping Guan.
\newblock Enlighten-your-voice: When multimodal meets zero-shot low-light image enhancement.
\newblock \emph{arXiv preprint arXiv:2312.10109}, 2023.

\bibitem[Zhang et~al.(2021)Zhang, Guo, Ma, Liu, and Zhang]{kind}
Yonghua Zhang, Xiaojie Guo, Jiayi Ma, Wei Liu, and Jiawan Zhang.
\newblock Beyond brightening low-light images.
\newblock \emph{International Journal of Computer Vision}, 129:\penalty0 1013--1037, 2021.

\bibitem[Zou et~al.(2022)Zou, Chen, Wu, Zhang, Xu, and Shao]{zou2022joint}
Wenbin Zou, Liang Chen, Yi~Wu, Yunchen Zhang, Yuxiang Xu, and Jun Shao.
\newblock Joint wavelet sub-bands guided network for single image super-resolution.
\newblock \emph{IEEE Transactions on Multimedia}, 25:\penalty0 4623--4637, 2022.

\bibitem[Zou et~al.(2024)Zou, Yu, Huang, and Zhao]{zou2024freqmamba}
Zhen Zou, Hu~Yu, Jie Huang, and Feng Zhao.
\newblock Freqmamba: Viewing mamba from a frequency perspective for image deraining.
\newblock In \emph{ACM Multimedia 2024}, 2024.

\end{thebibliography}
\bibliographystyle{iclr2025_conference}

\end{document}

\maketitle
\vspace{-10mm}
\section{implementation details}
\subsection{Datasets}
LOLv1~\cite{lolv1} is a classic low-light dataset containing images from various scenes under different lighting conditions, comprising 500 pairs of normal-light and low-light training images and 15 pairs of testing images. LOLv2-real~\cite{lolv2} is a dataset captured in real-world scenarios by varying ISO and exposure time, containing a total of 689 training image pairs and 100 testing image pairs. Both of these datasets are real-world datasets that include noise within the images. The images in the LOL series datasets are sized 400x600 pixels, and we follow the official train-test split ratio provided in the dataset. We utilized the combined 1189 images from the v1 and v2 training sets during training on the LOL series datasets, using only the unlabeled low-light data. In the testing phase, reference metrics were calculated using the normal-light data. 

SICE~\cite{sice} is a large-scale multi-exposure image dataset. The SICE dataset contains high-resolution multi-exposure image sequences that cover a diverse range of scenes. In the dataset, Multi-Exposure Fusion (MEF) and High Dynamic Range (HDR) techniques are used to reconstruct reference images. Through a detailed process from image capture to filtering and reference generation, 1,200 sequences were combined with 13 MEF/HDR algorithms, resulting in 15,600 fused images. After careful selection, 589 high-quality reference images and their corresponding sequences were retained for further use. We followed the same settings as those used in PairLie~\cite{pairlie} for training and dataset splitting. 

SIDD~\cite{sidd} is a real-world noisy dataset captured using smartphone cameras. For our experiments, we randomly selected low-light images from the SIDD Small dataset. The selected images were cropped into patches of size 1024x512, from which 1,500 images were used for training and 80 images for testing. Due to the dataset's inclusion of reference images captured under various camera settings, it is challenging to identify the most appropriate reference. Consequently, we utilized no-reference evaluation metrics to assess the enhancement performance.
\subsection{Training Details}
In the experiments, we employed the 'seed\_torch' function to set the random seeds for Python, NumPy, and PyTorch to "123." The loss function is defined as \(\mathcal{L} = \omega_R \mathcal{L}_R + \omega_L \mathcal{L}_L + \omega_{con} \mathcal{L}_{con} + \omega_{enh} \mathcal{L}_{enh}\), with the weights \(\omega_R\), \(\omega_L\), \(\omega_{con}\), and \(\omega_{enh}\) set to a ratio of 1:1:0.1:1. In \(\mathcal{L}_{enh}\), the ratio of the components controlling exposure to those controlling color balance is 1:0.5. During the preprocessing of the training data, we randomly sampled images from the initialized dataset and cropped them into patches of size 256×256 pixels. Additionally, we applied data augmentation techniques—specifically horizontal flipping, vertical flipping, and image rotation—to mitigate overfitting. Finally, the data was normalized to the range of (0, 1) before being fed into the deep network.

\section{DCT and IDCT}
As for frequency prior, we use channel-wise 2D DCT to convert the spatial-domain image $I$ into the frequency-domain counterpart $F$. Different spectral bands in the DCT domain encode different image visual attributes degradation representation analysis of input images: 
\begin{equation}\small
    F_{u,v}\!=\!\frac{2}{\sqrt{hw}}m(u)m(v)\sum\nolimits_{i=0}^{h-1}\sum\nolimits_{j=0}^{w-1}[I_{i,j} 
    \cos\frac{(2i+1)u\pi}{2h}\cos\frac{(2j+1)v\pi}{2w}], m(\gamma )\!=\!\left\{\begin{matrix} 
  \frac{1}{\sqrt{2} }, \gamma \!=\!0  \\  
  1,\gamma \!>\!0 
\end{matrix}\right.
\end{equation}

where the index $i$, $j$ denote the 2D coordinate in the spatial domain, while $u$, $v$ refer to the 2D coordinate in the DCT frequency domain.

By performing an inverse Discrete Cosine Transform (IDCT) on these filtered maps $F_{*}$, we obtain the corresponding spatial domain feature images:
\vspace{-2mm}
\begin{equation}\small
    C^*_{i,j}\!=\!\frac{2}{\sqrt{hw}}\sum\nolimits_{u=0}^{h-1}\sum\nolimits_{v=0}^{w-1}[m(u)m(v)F^*_{u,v} 
    \cos\frac{(2i+1)u\pi}{2h}\cos\frac{(2j+1)v\pi}{2w}],  m(\gamma )\!=\!\left\{\begin{matrix} 
  \frac{1}{\sqrt{2} }, \gamma \!=\!0  \\  
  1,\gamma \!>\!0 
\end{matrix}\right.
\end{equation}
\section{experiments}
\subsection{Comparison of cascading methods}
In the main text, we mention that most foundational frameworks for joint tasks employ multi-stage training and utilize a cascaded approach to handle each degradation task sequentially. To evaluate the performance difference between this approach and ours, we selected no-reference unsupervised denoising and enhancement methods for a cascaded implementation of this task. For unsupervised denoising, we employed the neighbor masking pipeline from Neighbor2neighbor~\cite{huang2021neighbor2neighbor}. Additionally, we paired two unsupervised low-light enhancement methods without explicit denoising design, EnlightenGAN~\cite{jiang2021enlightengan} and PairLIE~\cite{pairlie}, with the denoising method in a sequential inference. This resulted in four combinations arranged based on their processing order.

\begin{figure}[t] % 'h' 表示在此处插入图片
    \centering % 居中
    \includegraphics[width=\textwidth]{ICLR 2025 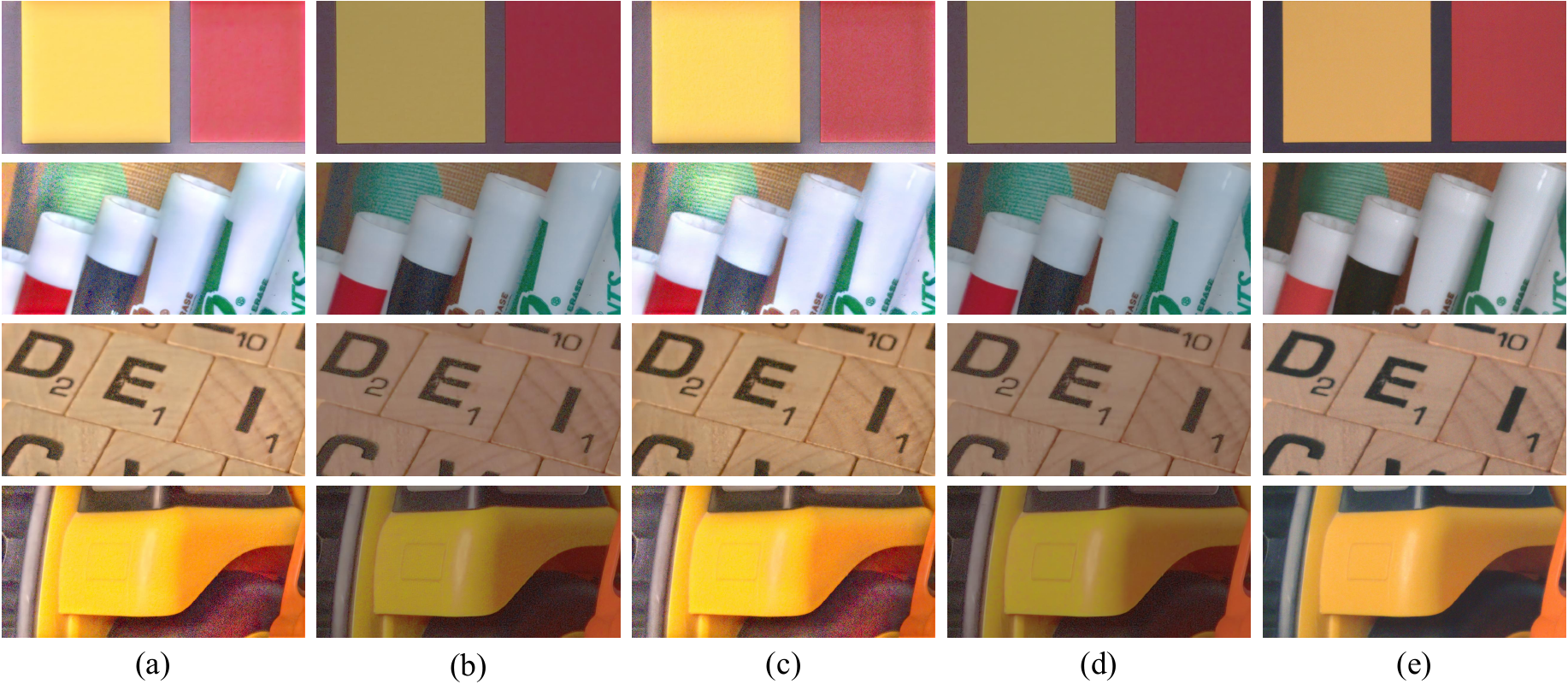} % 调整图片宽度和路径
    \vspace{-4mm}
    \caption{The visual results of the cascaded methods. (a) EnlightenGAN-Neighbor2neighbor. (b) PairLIE-Neighbor2neighbor. (c) Neighbor2neighbor-EnlightenGAN. (d) Neighbor2neighbor-PairLIE. (e)Ours.}  % 图片标题
    \vspace{-5mm}
    \label{fig:cascad} % 图片标签，用于引用
\end{figure}

The qualitative results of this comparative experiment are shown in Tab.~\ref{tab:cascad}, with corresponding visual results in Fig.~\ref{fig:cascad}. We observed that the image quality generated by these simple cascaded combinations tends to be worse than that of single-task methods (i.e., using only enhancement techniques). Specifically, local overexposure and the generation of more complex noise patterns are common. This is due to error accumulation during the sequential processing. Low-light enhancement methods often introduce nonlinear perturbations to the original noise, making the noise pattern more challenging to model and distinguish. Additionally, after denoising, the image's statistical properties (e.g., brightness, contrast, dynamic range) may shift, further complicating the dynamic range and leading the enhancement algorithm to overcompensate or underperform.
% \usepackage{multirow}
\begin{table}[t]\small
\centering

\caption{The cascaded ablation study on the low-light SIDD dataset presents no-reference quantitative metrics, where the top-performing method is highlighted in {\color[HTML]{FE0000} \textbf{red}} and the second-best method is marked in \color[HTML]{3166FF} \textbf{blue}.}
\vspace{-3mm}
\begin{tabular}{c|cc|cc|ccc}
\hline
\multirow{Methods} & \multicolumn{2}{c|}{Denoise-Enhance} & \multicolumn{2}{c|}{Enhance-Denoise} & \multicolumn{3}{c}{Single Method} \\
                         & EnlightenGAN        & PairLIE        & EnlightenGAN        & PairLIE        & EnlightenGAN  & PairLIE  & Ours   \\ \hline
BRISQUE\                  & 24.727              & 36.100         & 13.382              & 47.653         & 13.786        & {\color[HTML]{3166FF} \textbf{3.168}}    & {\color[HTML]{FE0000} \textbf{2.555}}  \\
CLIPIQA                  & 0.341               & 0.377          & 0.355               & 0.359          & {\color[HTML]{3166FF} \textbf{0.337}}         & 0.383    & {\color[HTML]{FE0000} \textbf{0.292}}  \\ \hline
\end{tabular}
\label{tab:cascad}
\vspace{-4mm}
\end{table}
\subsection{Generalization Evaluation}
\textbf{Testing on unsupervised datasets.} To evaluate the generalization capability of our proposed method, we selected SCI~\cite{SCI}, PairLIE~\cite{pairlie}, and RUAS~\cite{RUAS} as baseline methods. These approaches, along with our method, were applied to process datasets without ground truth(LIME~\cite{lime}, NPE~\cite{npe}, MEF~\cite{mef}, DICM~\cite{dicm} and VV~\cite{vv}) using models trained on the LOL dataset. For a fair comparison, all methods were trained for 100 epochs. We employed two widely-used no-reference quality assessment metrics, NIQE and BRISQUE, as benchmarks to qualitatively evaluate the performance of image enhancement.
\vspace{2mm}
\begin{figure}[h] % 'h' 表示在此处插入图片
    \centering % 居中
    \includegraphics[width=\textwidth]{ICLR 2025 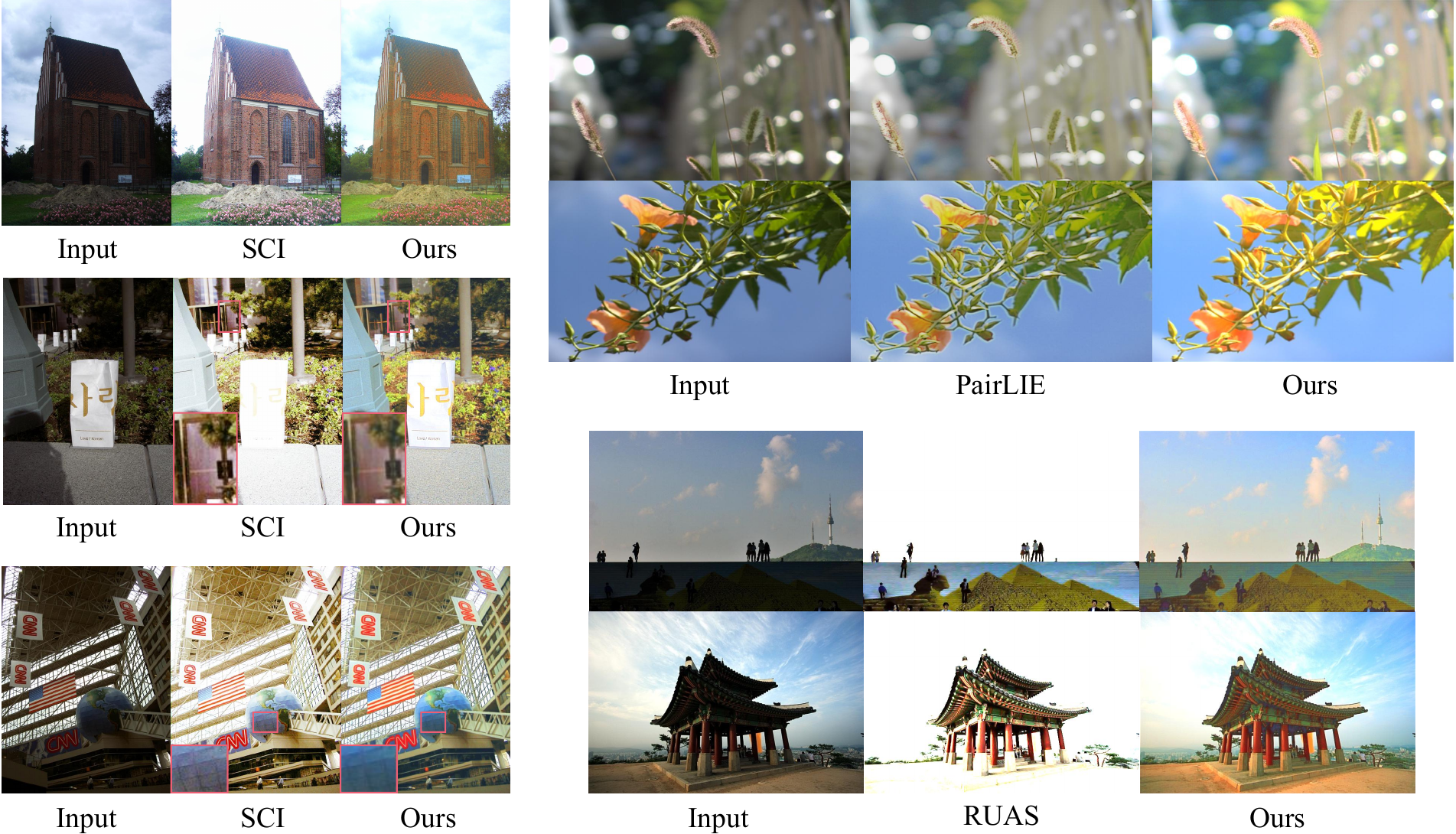} % 调整图片宽度和路径
    \vspace{-4mm}
    \caption{Qualitative results on the unsupervised dataset.} % 图片标题
    \vspace{-4mm}
    \label{fig:unsuper} % 图片标签，用于引用
\end{figure}
\begin{table}[h]
\centering
\caption{Quantitative comparisons on the unsupervised dataset LIME, NPE and MEF, where the top-performing method is highlighted in {\color[HTML]{FE0000} \textbf{red}} and the second-best method is marked in \color[HTML]{3166FF} \textbf{blue}.}
\vspace{-2mm}
\begin{tabular}{c|cc|cc|cc}
\hline
Dataset                         & \multicolumn{2}{c|}{LIME}                                                      & \multicolumn{2}{c|}{NPE}                                                       & \multicolumn{2}{c}{MEF}                                                        \\
Method  & NIQE$\downarrow$                                  & BRISQUE$\downarrow$                                & NIQE$\downarrow$                                  & BRISQUE$\downarrow$                                & NIQE$\downarrow$                                  & BRISQUE$\downarrow$                                \\ \hline
RUAS    & 5.376                                 & 28.937                                 & 7.060                                 & 49.594                                 & 5.423                                 & 33.817                                 \\
PairLIE & 4.569                                 & 23.699                                 & {\color[HTML]{3166FF} \textbf{4.137}} & {\color[HTML]{3166FF} \textbf{21.528}} & 4.288                                 & 28.388                                 \\
SCI     & {\color[HTML]{3166FF} \textbf{4.182}} & {\color[HTML]{3166FF} \textbf{19.701}} & 4.473                                 & 27.657                                 & {\color[HTML]{FE0000} \textbf{3.634}} & {\color[HTML]{FE0000} \textbf{14.399}} \\
Ours                            & {\color[HTML]{FE0000} \textbf{4.109}} & {\color[HTML]{FE0000} \textbf{16.382}} & {\color[HTML]{FE0000} \textbf{3.802}} & {\color[HTML]{FE0000} \textbf{17.140}} & {\color[HTML]{3166FF} \textbf{3.758}} & {\color[HTML]{3166FF} \textbf{18.997}} \\ \hline
\end{tabular}
\label{tab:lime}
\end{table}
\vspace{-4mm}
\begin{table}[h]
    \centering
    \caption{Quantitative comparisons on the unsupervised dataset DICM and VV, where the top-performing method is highlighted in {\color[HTML]{FE0000} \textbf{red}} and the second-best method is marked in \color[HTML]{3166FF} \textbf{blue}.}
    \vspace{-2mm}
\begin{tabular}{c|ll|ll}
\hline
Dataset                         & \multicolumn{2}{c|}{DICM}                                                      & \multicolumn{2}{c}{VV}                                                        \\
Method  & \multicolumn{1}{c}{NIQE}              & \multicolumn{1}{c|}{BRISQUE}           & \multicolumn{1}{c}{NIQE}              & \multicolumn{1}{c}{BRISQUE}           \\ \hline
RUAS    & 7.052                                 & 46.522                                 & 5,297                                 & 51.085                                 \\
PairLIE & {\color[HTML]{3166FF} \textbf{4.064}} & 30.833                                 & {\color[HTML]{3166FF} \textbf{3.648}} & {\color[HTML]{333333} 31.213}          \\
SCI     & {\color[HTML]{333333} 4.073}          & {\color[HTML]{3166FF} \textbf{27.706}} & {\color[HTML]{FE0000} \textbf{2.934}} & {\color[HTML]{FE0000} \textbf{21.431}} \\
Ours                            & {\color[HTML]{FE0000} \textbf{3.859}} & {\color[HTML]{FE0000} \textbf{26.592}} & {\color[HTML]{333333} 3.748}          & {\color[HTML]{3166FF} \textbf{29.701}} \\ \hline
\end{tabular}
\label{tab:vv}
\vspace{-3mm}
\end{table}

The experimental results, as shown in the Fig.\ref{fig:unsuper}, Tab.\ref{tab:lime} and\ref{tab:vv}, demonstrate that our method achieves superior performance in noise suppression, exposure stability, and natural color restoration. Compared to SCI, our approach generates fewer artifacts and noise in local details (e.g., the boxed area on the left of Fig.\ref{fig:unsuper}). Compared to PairLIE, our method produces richer color gradations and aesthetically pleasing details, with smoother handling of light and shadows. In comparison with RUAS, our approach exhibits better generalization under the exposure conditions of the unsupervised dataset, adaptively enhancing dark regions while avoiding overexposure in originally bright areas.

\textbf{Testing on overexposed images.}In addition, we explored the generalization capability of the proposed method on overexposed images. Specifically, we employed the model trained for 100 epochs on the LOL dataset to process overexposed images. The selected test dataset consists of the overexposed samples from the test set of the Exposure-Error Dataset in \cite{exposure_correction}.

\begin{figure}[h] % 'h' 表示在此处插入图片
    \centering % 居中
    \includegraphics[width=\textwidth]{ICLR 2025 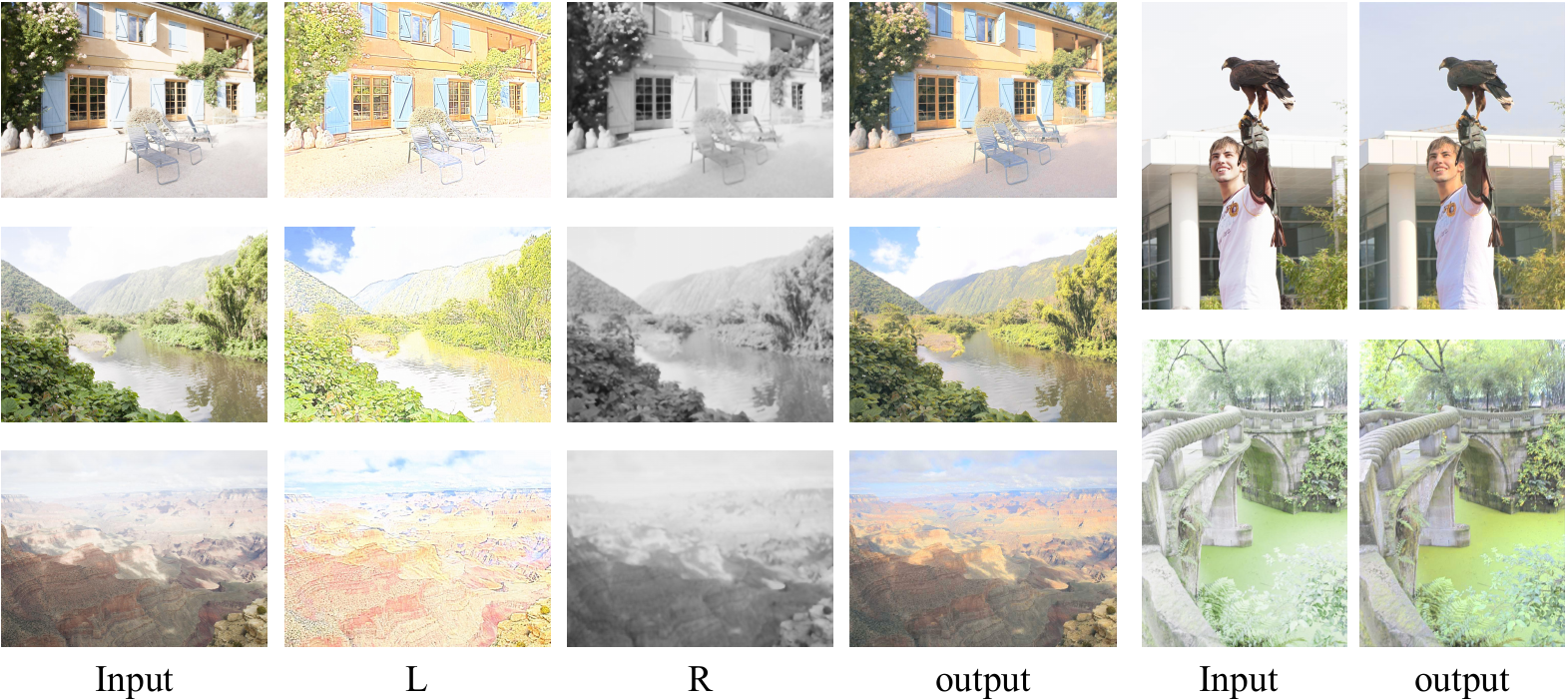} % 调整图片宽度和路径
    \vspace{-2mm}
    \caption{Visualization results on the overexposed dataset.} % 图片标题
    \label{fig:exp} % 图片标签，用于引用
    \vspace{-2mm}
\end{figure}

As illustrated in the experimental results Fig.\ref{fig:exp}, the proposed method demonstrates generalization to overexposed scenarios and achieves accurate Retinex decomposition. However, certain limitations are observed. Since our adaptive illumination adjustment module (LCNet) is trained on low-light datasets, domain discrepancies between the two tasks often result in lower output values from LCNet. This causes the enhanced image to maintain relatively high exposure levels. To address this issue, we adopt enhancement measures inspired by PairLIE, leveraging traditional techniques to adjust the decomposed illumination map. Specifically, dynamic range adjustment is employed. The final output exhibits richer color gradations and exposure levels closer to human visual perception, compared to the input image.
\vspace{-4mm}
\subsection{ablation study}
\begin{table}[]
\vspace{-4mm}
\centering
\caption{The impact of the loss function on model performance evaluated on the LOLv1 and LOLv2 dataset. The data ranked first is highlighted in \color[HTML]{FE0000} \textbf{red}.}
\vspace{-3mm}
\begin{tabular}{ccccccc}
\hline
Dataset      & \multicolumn{3}{c}{LOLv1}                                                                                             & \multicolumn{3}{c}{LOLv2}                                                                                             \\
Setting      & PSNR                                  & SSIM                                  & LPIPS                                 & PSNR                                  & SSIM                                  & LPIPS                                 \\ \hline
w/o $\mathcal{L}_R$            & 18.93                                 & 0.713                                 & 0.266                                 & 19.40                                 & 0.723                                 & 0.290                                 \\
w/o $\mathcal{L}_L$            & 12.61                                 & 0.542                                 & 0.725                                 & 12.05                                 & 0.492                                 & 0.705                                 \\
w/o $\mathcal{L}_{enh}$            & 7.30                                  & 0.130                                 & 0.720                                 & 9.12                                  & 0.140                                 & 0.699                                 \\
w/o $\mathcal{L}_{con}$            & 19.49                                 & 0.744                                 & 0.279                                 & 19.72                                 & 0.759                                 & 0.277                                 \\
Full Version & {\color[HTML]{FE0000} \textbf{19.80}} & {\color[HTML]{FE0000} \textbf{0.750}} & {\color[HTML]{FE0000} \textbf{0.253}} & {\color[HTML]{FE0000} \textbf{20.22}} & {\color[HTML]{FE0000} \textbf{0.793}} & {\color[HTML]{FE0000} \textbf{0.266}} \\ \hline
\end{tabular}
\vspace{-4mm}
\end{table}
\textbf{The impact of the gamma coefficient control factor.} We analyzed the impact of the gamma coefficient control factor \(\sigma\) on image restoration performance. Initially, as \(\sigma\) increases, PSNR exhibits an upward trend, while LPIPS shows a downward trend. The optimal denoising and enhancement effects occur at \(\sigma = 1.5\), where both metrics reach their respective optimal ranges. The variation in SSIM follows a similar trend, peaking at \(\sigma = 1.8\). This indicates that \(\sigma = 1.5\) represents the optimal point for overall image quality, with minimal distortion and superior perceptual quality, while SSIM reaches its highest value at \(\sigma = 1.8\), albeit with negligible overall variation. Thus, our results suggest that \(\sigma = 1.5\) is the optimal parameter for image restoration.

Furthermore, we investigated the restoration outcomes across three variable intervals. The findings indicate that when \(\sigma\) is subjected to random sampling within a specified interval, the resultant restorations significantly outperform those achieved with a fixed factor at the midpoint of the interval. We propose that this random sampling approach mitigates the model's tendency to learn the identity transformation imposed by gamma nonlinear enhancement, thereby bolstering the robustness of the retinal decomposition process. Accordingly, we identified the optimal interval \((1.3, 1.7)\) for our experimental configuration.
\begin{table}[t]
    \centering
    \begin{minipage}{0.5\linewidth} % 设置第一个表格宽度
        \centering
        \begin{tabular}{cccc}

\hline
\renewcommand{\arraystretch}{2}
$\sigma$ & PSNR  & SSIM  & LPIPS \\ \hline
1.2   & 18.95 & 0.735 & 0.279 \\
1.3   & 19.59 & 0.737 & 0.256 \\
1.4   & 19.68 & 0.738 & 0.255 \\
1.5   & {\color[HTML]{FE0000} \textbf{19.74}} & 0.740 & {\color[HTML]{FE0000} \textbf{0.254}} \\
1.6   & 19.68 & 0.741 & 0.255 \\
1.7   & 19.70 & 0.741 & 0.257 \\
1.8   & 19.65 & {\color[HTML]{FE0000} \textbf{0.742}} & 0.257 \\
1.9   & 19.55 & 0.737 & 0.260 \\\hline
1.2-1.6   & 19.69 & 0.740 & 0.255 \\
1.3-1.7   & {\color[HTML]{FE0000} \textbf{19.80}} & {\color[HTML]{FE0000} \textbf{0.750}} & {\color[HTML]{FE0000} \textbf{0.253}} \\
1.4-1.8   & 19.71 & 0.742 & 0.256 \\\hline
\end{tabular}
        \caption{The impact of the regulation factor \( \sigma \) on model performance evaluated on the LOLv1 dataset. The data ranked first is highlighted in \color[HTML]{FE0000} \textbf{red}.} % 局部表格标题
    \end{minipage}
    \hfill % 增加两表格间的间距
    \begin{minipage}{0.45\linewidth} % 设置第二个表格宽度
        \centering
        \begin{tabular}{ccc}
\hline
Method     & Flops(G)                                & Time(ms)                                 \\ \hline
Retinexnet & 14.23                                & 7.37                                 \\
LLFormer   & 3.46                                 & 51.99                                \\
SNR-aware  & 6.97                                 & 19.89                                 \\ \hline
Zero-DCE   & 1.30                                 & {\color[HTML]{FE0000} \textbf{1.39}} \\
RUAS       & 0.05                                 & 3.72                                 \\
Clip-LIT   & 4.56                                 & 2.10                                 \\
PairLIE    & 5.59                                 & 1.70                                 \\
SCI        & {\color[HTML]{FE0000} \textbf{0.01}} & 1.52                                 \\
ours       & 5.10                                 & 16.56                                \\ \hline
\end{tabular}
        \caption{Comparison of Computational Complexity and Runtime Efficiency. The data ranked first is highlighted in \color[HTML]{FE0000} \textbf{red}.} % 局部表格标题
    \end{minipage}
    \label{tab:two-tables}
    \vspace{-5mm}
\end{table}

\begin{figure}[h] % 'h' 表示在此处插入图片
    \centering % 居中
    \includegraphics[width=\textwidth]{ICLR 2025 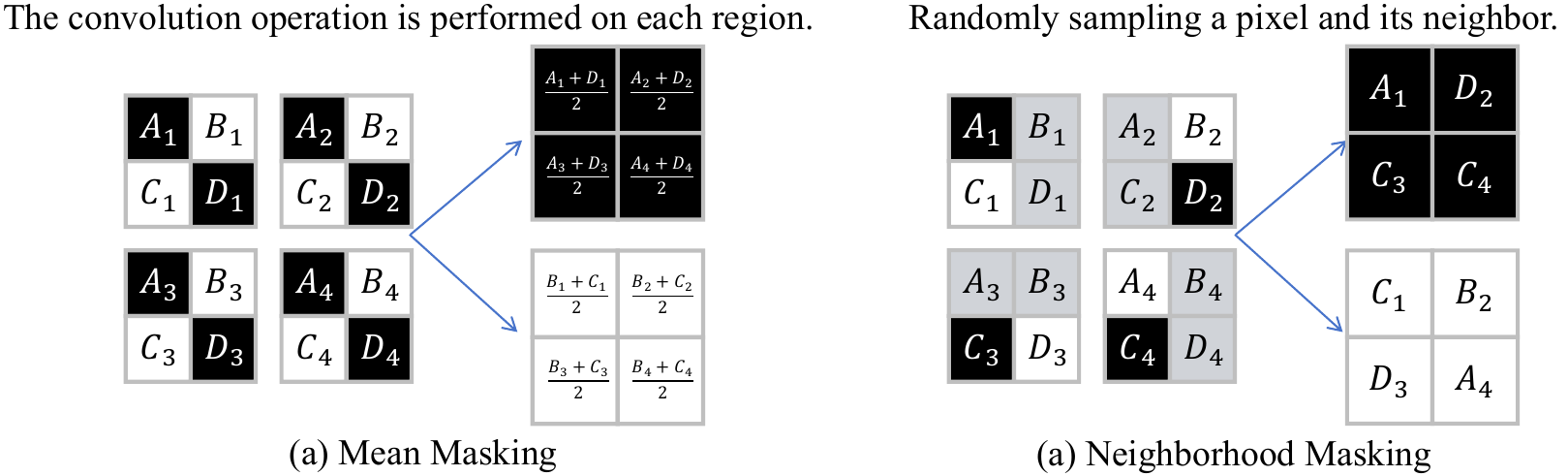} % 调整图片宽度和路径
    \vspace{-4mm}
    \caption{The schematic diagram illustrating the mechanisms of two mask strategies.} % 图片标题
    \vspace{-5mm}
    \label{fig:mask} % 图片标签，用于引用
\end{figure}

\textbf{the impact of different masking strategies.} Additionally, we analyzed the impact of different masking strategies on the performance of the joint framework. We selected the Neighborhood Masking proposed by Neighbor2Neighbor~\cite{huang2021neighbor2neighbor} and the Mean Masking introduced by ZS-N2N~\cite{zsn2n}, as illustrated in Fig.~\ref{fig:mask}. We have supplemented our work with ablation studies on masking strategies. All experiments were conducted under the same settings. We referred to the horizontal and vertical masking strategies from Noise2Fast~\cite{noise2fast}. The specific steps are as follows:

Noise2Fast-H: The image is divided into patches of size 2×1 pixels along the height direction. The top and bottom pixels within each patch are placed into the corresponding positions of two sub-images, respectively. After masking, the sub-image dimensions are \(H/2 \times W\), containing 50\% of the pixel information.  

Noise2Fast-W: The image is divided into patches of size 1×2 pixels along the width direction. The left and right pixels within each patch are placed into the corresponding positions of two sub-images, respectively. After masking, the sub-image dimensions are \(H \times W/2\), containing 50\% of the pixel information.  

Mean Masking: The image is divided into patches of size 2×2 pixels. The average of the pixels along the two diagonals is computed and placed into the corresponding positions of two sub-images. After masking, the sub-image dimensions are \(H/2 \times W/2\), containing 50\% of the pixel information.  

All three methods employ deterministic masking strategies, whereas Neighbor-masking adopts a stochastic masking approach. The experimental results are presented in Tab.~\ref{tab:mask}. 

\begin{table}[h]\small
\centering
\caption{The ablation study on the impact of different masking strategies on the joint framework. The data ranked first is highlighted in \color[HTML]{FE0000} \textbf{red}.}
\vspace{-3mm}
\begin{tabular}{c|ccc|ccc}
\hline
Dataset               & \multicolumn{3}{c|}{LOLv1}                                                                                            & \multicolumn{3}{c}{LOLv2}                                                                                             \\
Masking Method       & PSNR                                  & SSIM                                  & LPIPS                                 & PSNR                                  & SSIM                                  & LPIPS                                 \\ \hline
Mean Masking         & 18.70                                 & {\color[HTML]{FE0000} \textbf{0.758}} & 0.258                                 & 20.17                                 & 0.787                                 & 0.263                                 \\
Noise2Fast-H         & 18.99                                 & 0.736 & 0.260                                 & 19.94                                 & 0.777                                 & 0.261                                 \\
Noise2Fast-W         & 19.05                                 & 0.744 & 0.253                                 & 19.80                                 & 0.783                                 & 0.267                                 \\
Neighborhood Masking & {\color[HTML]{FE0000} \textbf{19.80}} & 0.750                                 & {\color[HTML]{FE0000} \textbf{0.253}} & {\color[HTML]{FE0000} \textbf{20.22}} & {\color[HTML]{FE0000} \textbf{0.793}} & {\color[HTML]{FE0000} \textbf{0.266}} \\ \hline
\end{tabular}
\label{tab:mask}
\vspace{-3mm}
\end{table}
Based on the analysis of the results above, despite the fact that the other three methods mask a smaller percentage of pixels during sampling, they fail to outperform Neighbor-masking. We attribute this outcome to the following two reasons: 1) Reduced sample diversity due to deterministic sampling: Deterministic sampling limits the variability of the training data. For instance, the other three masking strategies produce only one possible image pair per sampling, whereas Neighbor-masking generates 4×2=8 possible pairs. 2) Impact of noise correlation in real-image denoising tasks: Noise in neighboring pixels tends to exhibit correlations, which can negatively affect masked denoising. Increasing the masking ratio, thus enlarging the distance between visible pixels, helps mitigate this correlation to some extent.  

Although intuitively, a higher masking ratio may result in the loss of local details and over-smoothing, the unique nature of image denoising tasks sets it apart from self-supervised image compression. Specifically, it requires additional consideration of noise correlation. After comprehensive evaluation, we opted for the current strategy. In future work, we plan to further explore alternative masking approaches.
\subsection{comparison on LSRW Dataset}
\begin{figure}[h] % 'h' 表示在此处插入图片
    \centering % 居中
    \includegraphics[width=\textwidth]{ICLR 2025 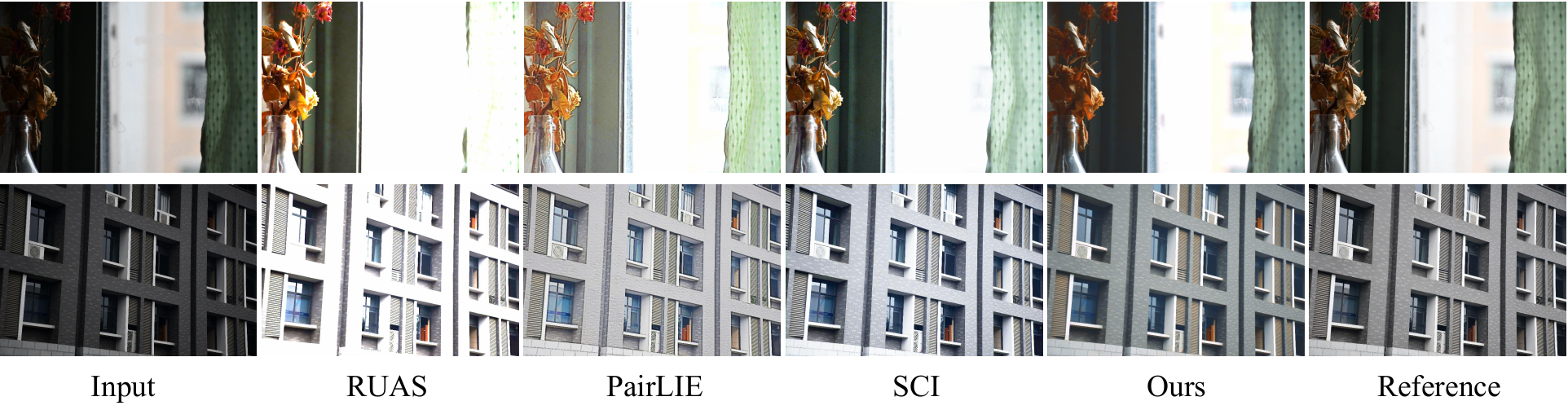} % 调整图片宽度和路径
    \caption{The visualization results on LSRW Dataset.} % 图片标题
    \label{fig:lsrw} % 图片标签，用于引用
\end{figure}

To further validate the reliability and stability of our proposed method, we conducted additional comparative experiments on the real-world LSRW dataset. The LSRW dataset consists of two subsets: Huawei smartphone and Nikon camera. We performed separate training and testing on each subset, using PSNR and SSIM as quantitative evaluation metrics. The experimental results are presented in Tab.\ref{tab:lsrw} and Fig.\ref{fig:lsrw}.
\begin{table}[]
\centering
\caption{uantitative comparisons on the real-world dataset LSRW, where the top-performing method is highlighted in {\color[HTML]{FE0000} \textbf{red}}.}
\vspace{-3mm}
\begin{tabular}{c|cc|cc}
\hline
Dataset & \multicolumn{2}{c|}{LSRW-Huawei}                                              & \multicolumn{2}{c}{LSRW-Nikon}                                                \\
Method  & PSNR                                  & SSIM                                  & PSNR                                  & SSIM                                  \\ \hline
RUAS    & 15.74                                 & 0.498                                 & 12.21                                 & 0.439                                 \\
PairLIE & {\color[HTML]{FE0000} \textbf{18.99}} & 0.550                                 & 15.52                                 & 0.427                                 \\
SCI     & 15.70                                 & 0.428                                 & 14.65                                 & 0.407                                 \\
Ours    & 18.94                                 & {\color[HTML]{FE0000} \textbf{0.558}} & {\color[HTML]{FE0000} \textbf{17.61}} & {\color[HTML]{FE0000} \textbf{0.493}} \\ \hline
\end{tabular}
\label{tab:lsrw}
\vspace{-3mm}
\end{table}

Compared to two other zero-reference methods, SCI and RUAS, as well as the unsupervised method PairLIE, our approach achieves superior results in both quantitative metrics and visual outcomes. From a visual perspective, RUAS and SCI suffer from the absence of adaptive illumination design, leading to overexposure or underexposure under uneven low-light conditions. In contrast, PairLIE tends to produce artifacts and excessively enhanced edges. Our method delivers the best perceptual results, addressing these challenges effectively.
\section{The supplementary visual results}
\begin{figure}[h] % 'h' 表示在此处插入图片
    \centering % 居中
    \includegraphics[width=\textwidth]{ICLR 2025 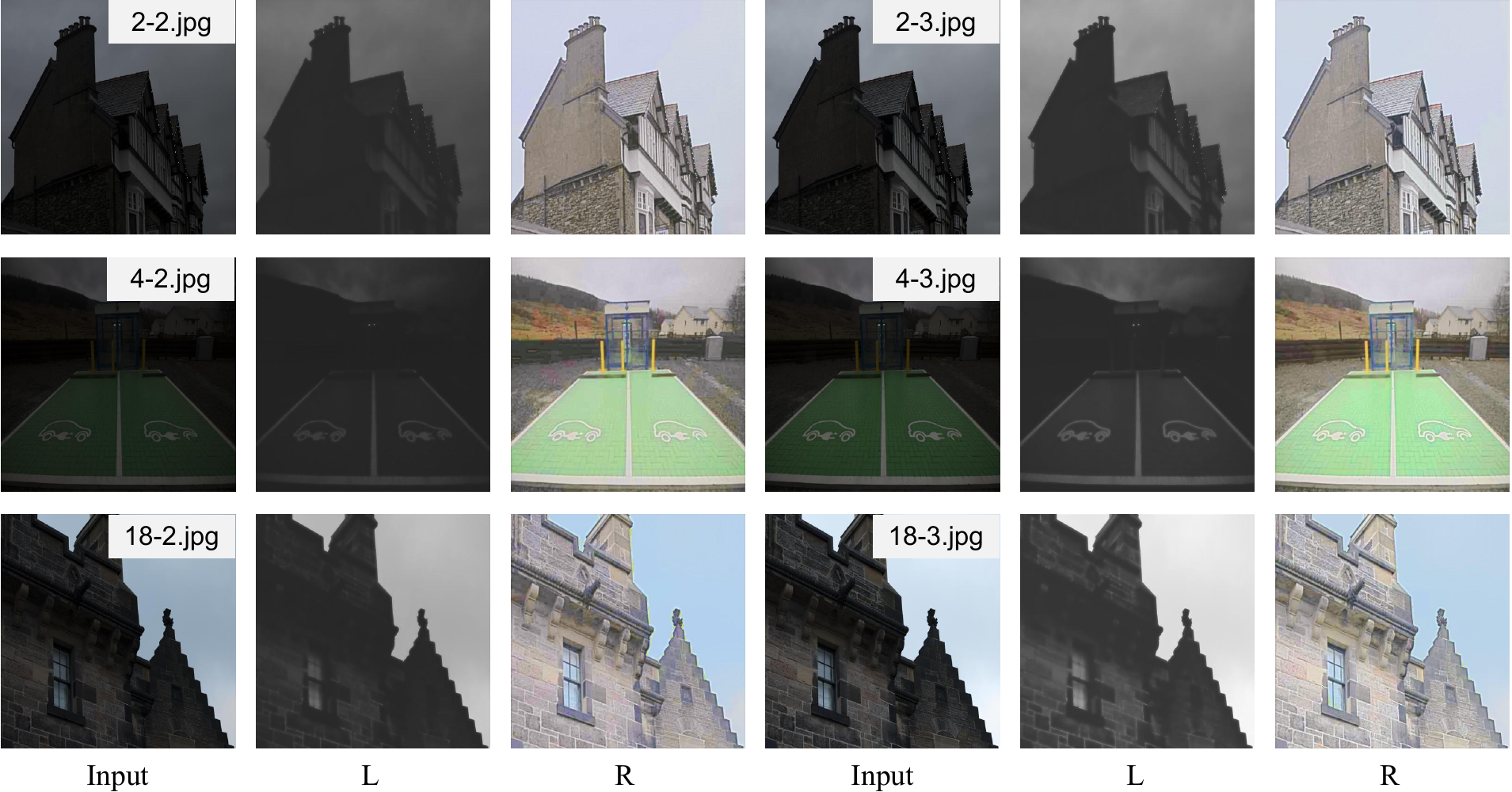} % 调整图片宽度和路径
    \caption{Retinex decomposition visual supplement results.} % 图片标题
    \label{fig:decompose} % 图片标签，用于引用
\end{figure}

\begin{figure}[h]
    \centering
    \begin{subfigure}{\linewidth}
        \centering
        \includegraphics[width=\linewidth]{ICLR 2025 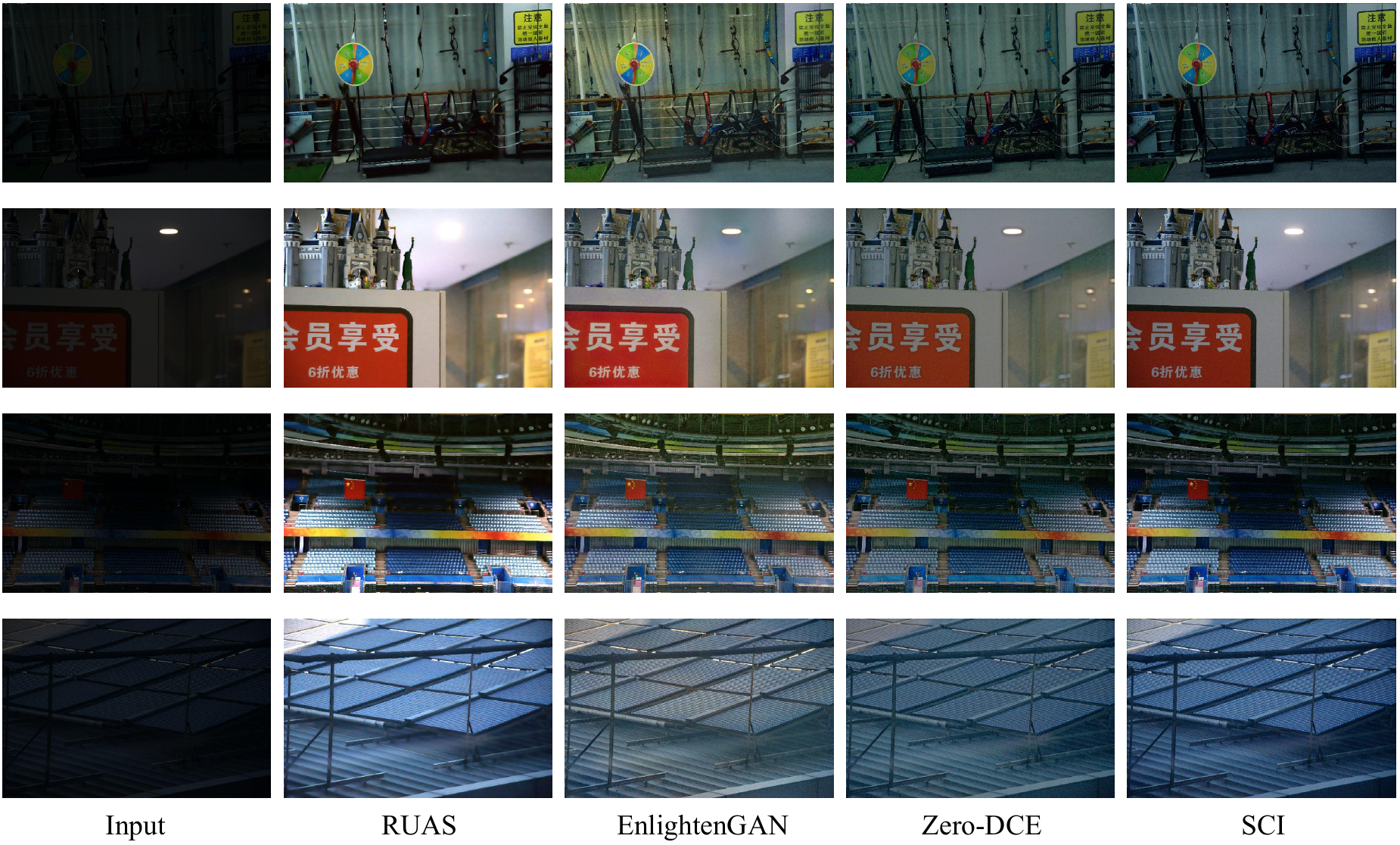}
        \label{fig:image1}
    \end{subfigure}
    \vspace{2mm}
 % 控制两张图片之间的垂直间距

    \begin{subfigure}{\linewidth}
        \centering
        \includegraphics[width=\linewidth]{ICLR 2025 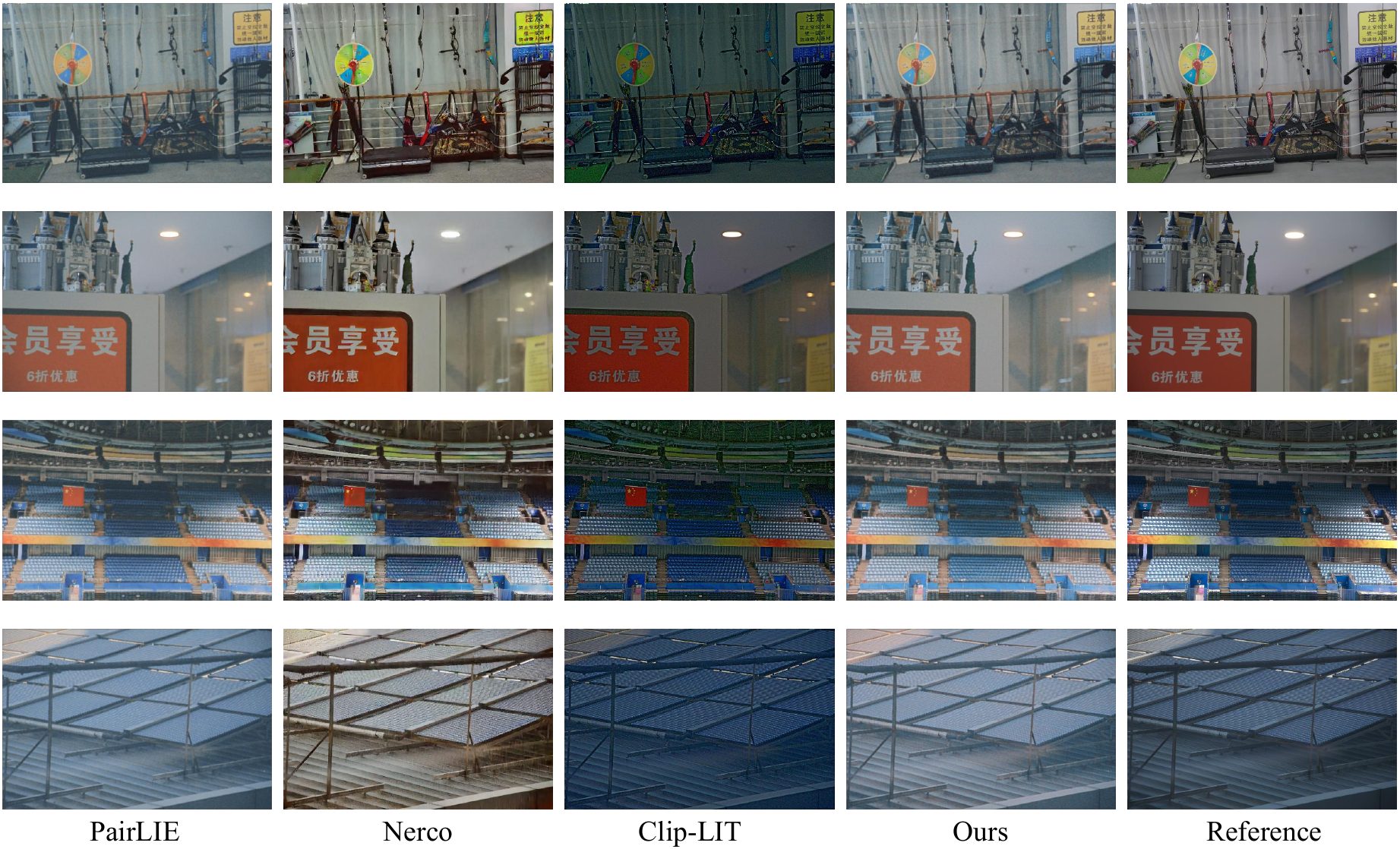}
        \label{fig:image2}
    \end{subfigure}

    \caption{Additional qualitative results from comparative experiments on the LOL dataset.}
    \label{fig:lol_sup}
    \vspace{10mm}
\end{figure}

\begin{figure}[h]
    \centering
    \begin{subfigure}{\linewidth}
        \centering
        \includegraphics[width=\linewidth]{ICLR 2025 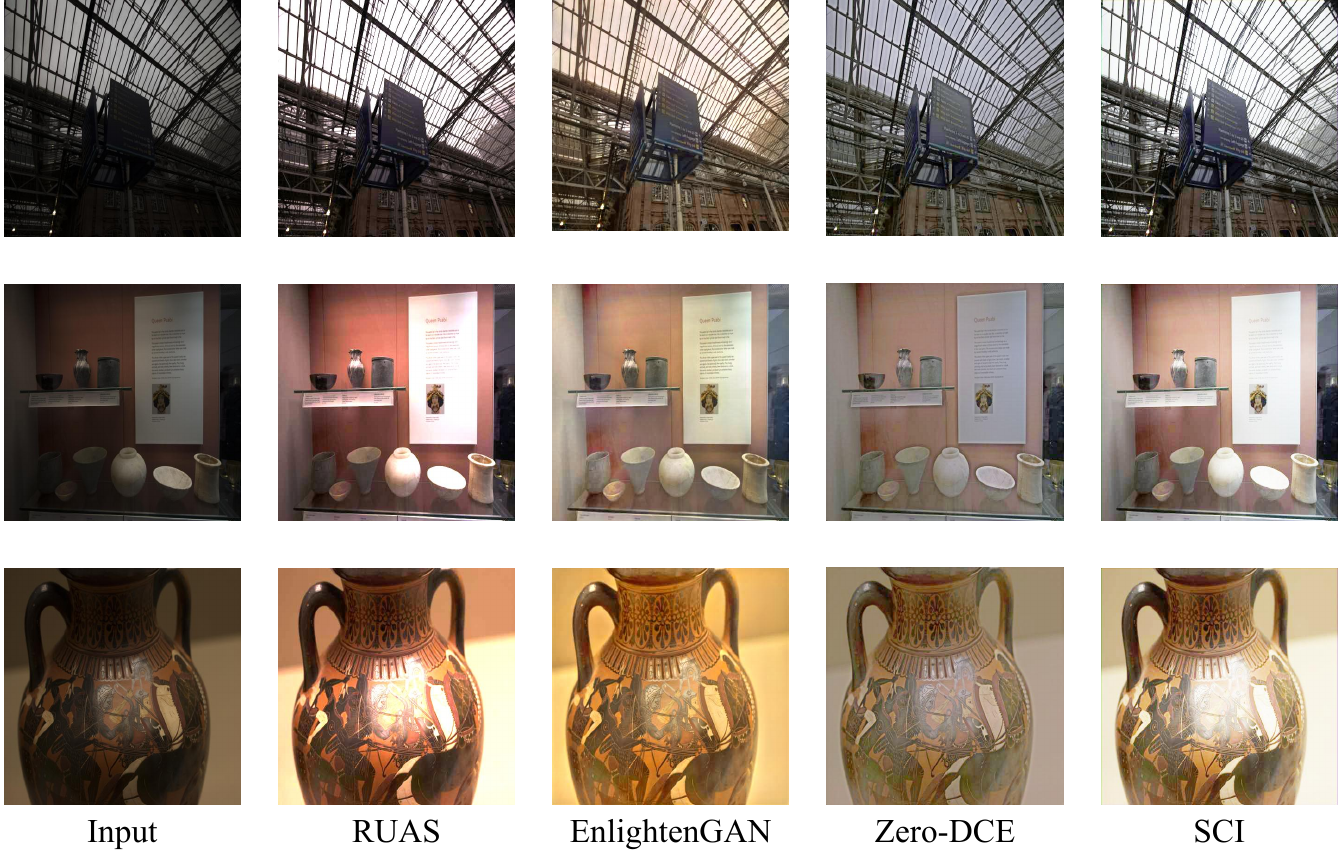}
        \label{fig:image1}
    \end{subfigure}
 % 控制两张图片之间的垂直间距

    \begin{subfigure}{\linewidth}
        \centering
        \includegraphics[width=\linewidth]{ICLR 2025 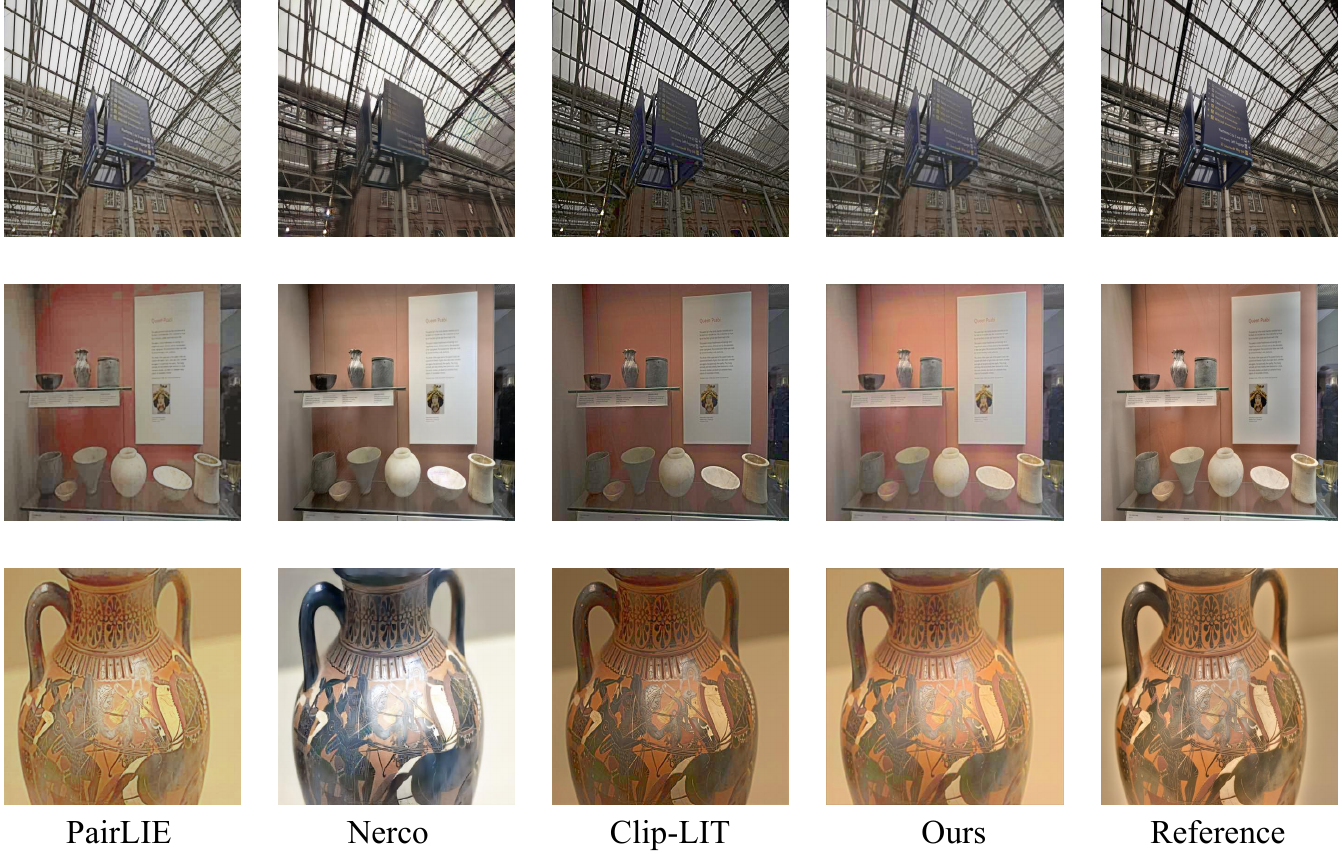}
        \label{fig:image2}
    \end{subfigure}

    \caption{Additional qualitative results from comparative experiments on the SICE dataset.}
    \label{fig:sice_sup}
    
\end{figure}

\begin{figure}[h]
    \centering
    \begin{subfigure}{\linewidth}
        \centering
        \includegraphics[width=\linewidth]{ICLR 2025 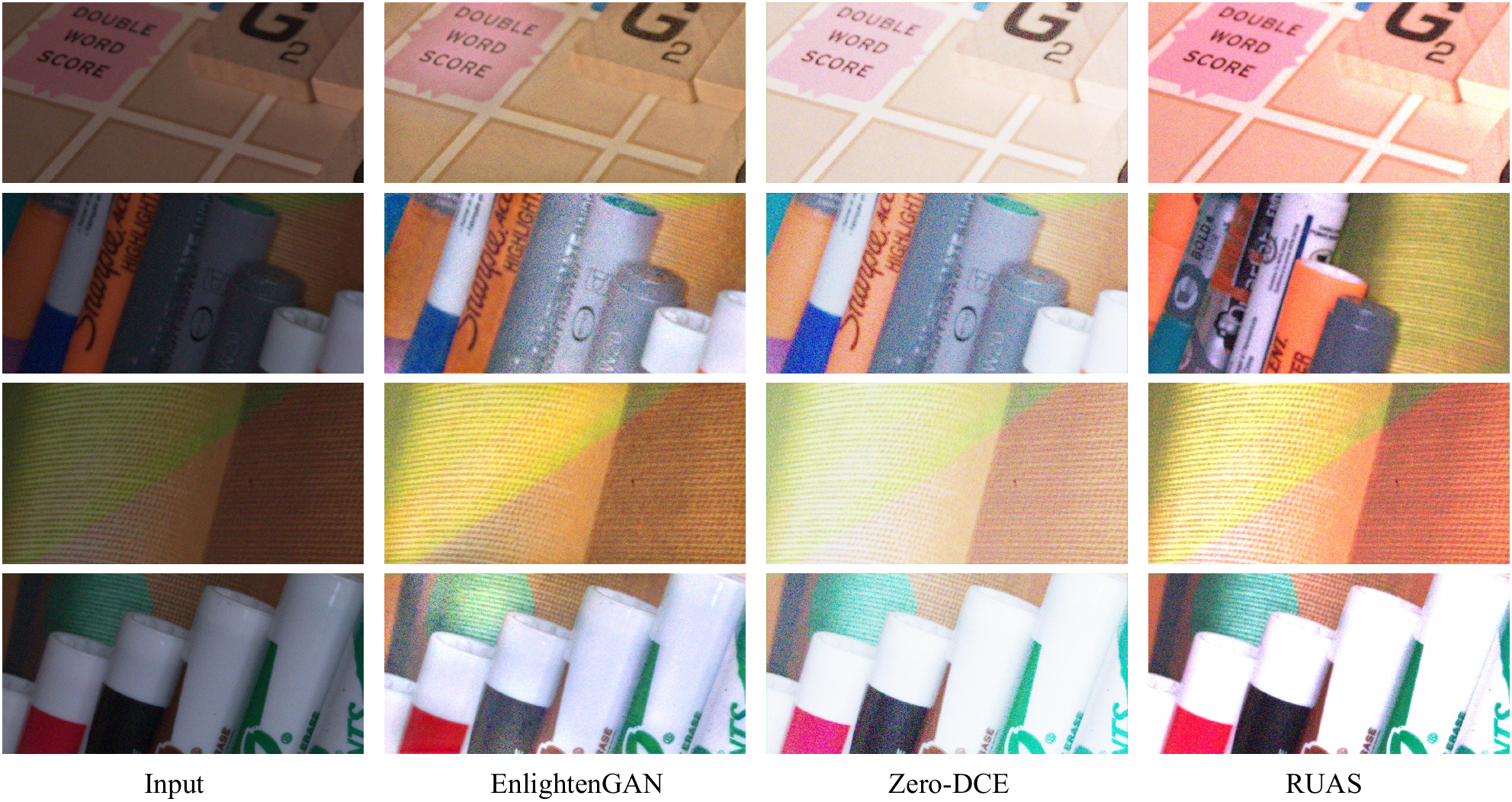}
        \label{fig:image1}
    \end{subfigure}
 % 控制两张图片之间的垂直间距

    \begin{subfigure}{\linewidth}
        \centering
        \includegraphics[width=\linewidth]{ICLR 2025 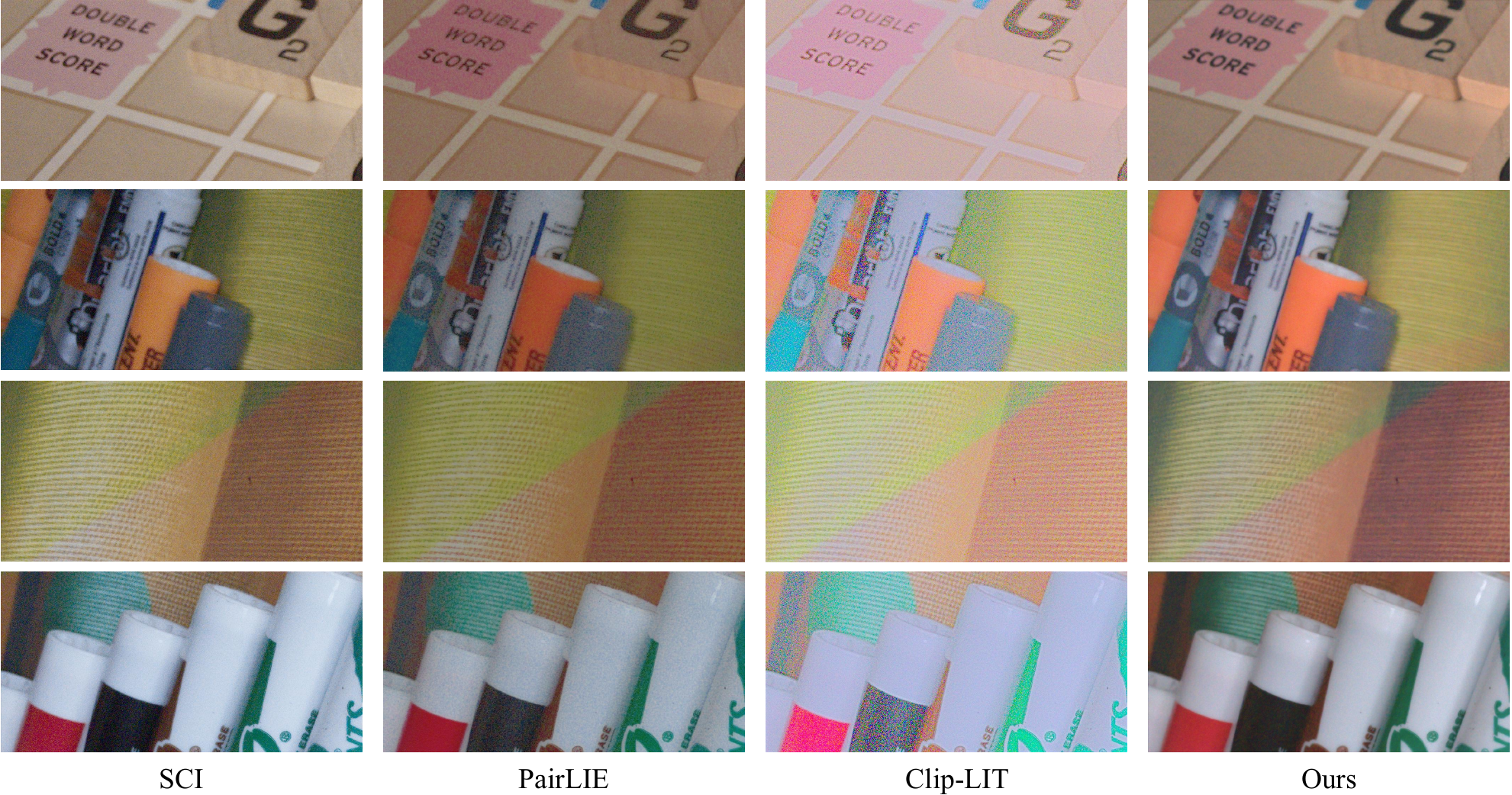}
        \label{fig:image2}
    \end{subfigure}

    \caption{Additional qualitative results from comparative experiments on the sidd dataset.}
    \label{fig:sidd_Sup}
\end{figure}

Here, we provide some supplementary visual results. Fig.\ref{fig:decompose} illustrates the Retinex decomposition visualization, while Fig.\ref{fig:lol_sup}, \ref{fig:sice_sup}, \ref{fig:sidd_Sup} present the results on the LOL, SICE, and SIDD datasets, respectively.
\begin{figure}[h] % 'h' 表示在此处插入图片
    \centering % 居中
    \includegraphics[width=13cm]{ICLR 2025 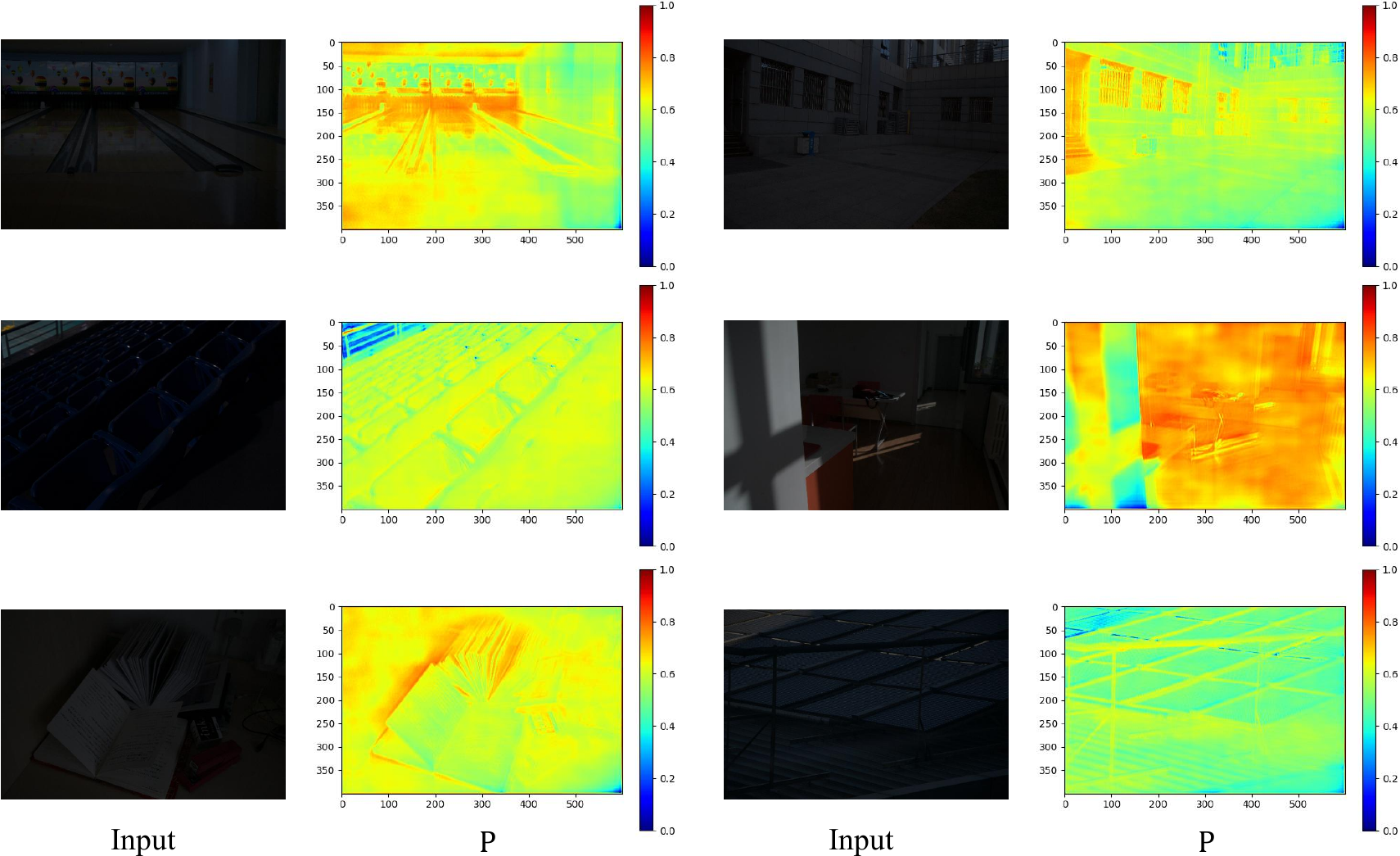} % 调整图片宽度和路径
    \caption{The visualization results of the degradation representation \(P\) reveal that \(P\) focuses its attention predominantly on the darker regions of the input image and effectively captures certain semantic information.} % 图片标题
    \label{fig:exp} % 图片标签，用于引用
\end{figure}
\clearpage
\begin{figure}[h]
    \centering
    \includegraphics[width=\linewidth]{ICLR 2025 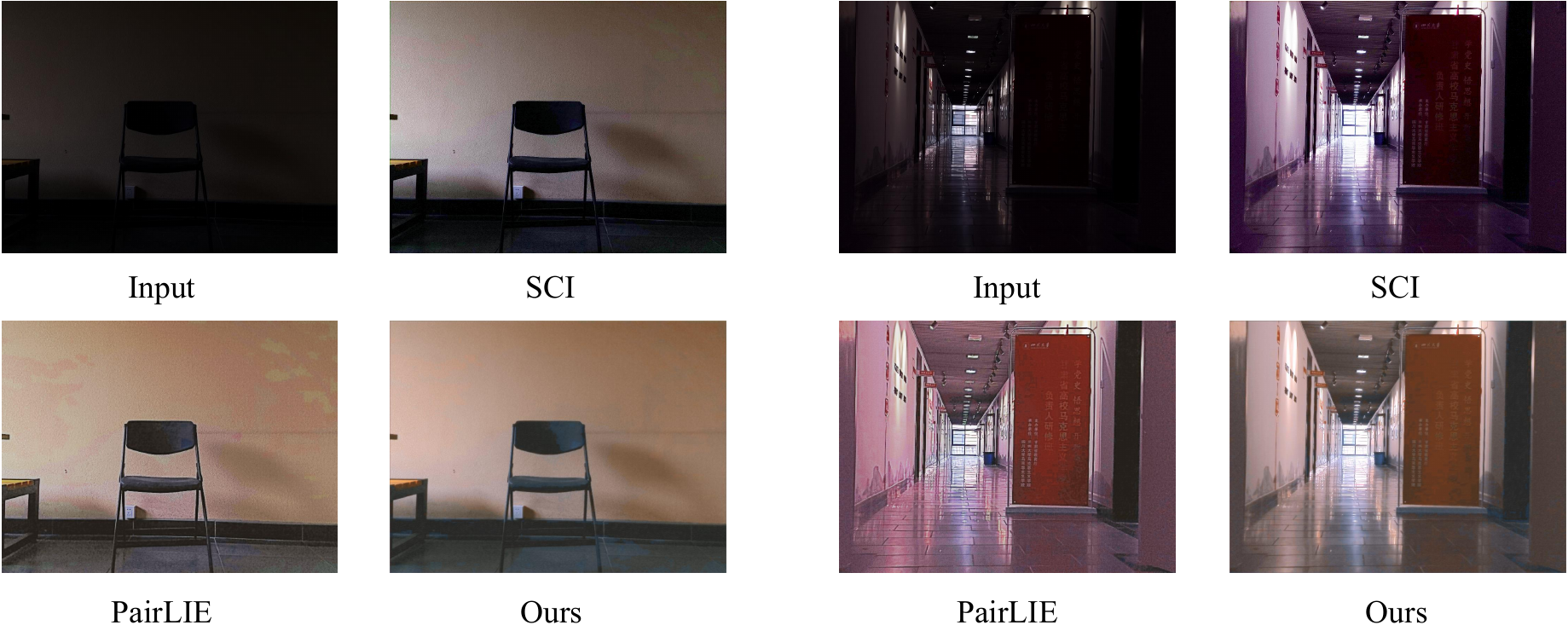}
    \caption{Additional qualitative results in edge situation with color offset and severe artifacts.}
    \label{fig:edge}
\end{figure}
\begin{figure}[h]
    \centering
    \includegraphics[width=\linewidth]{ICLR 2025 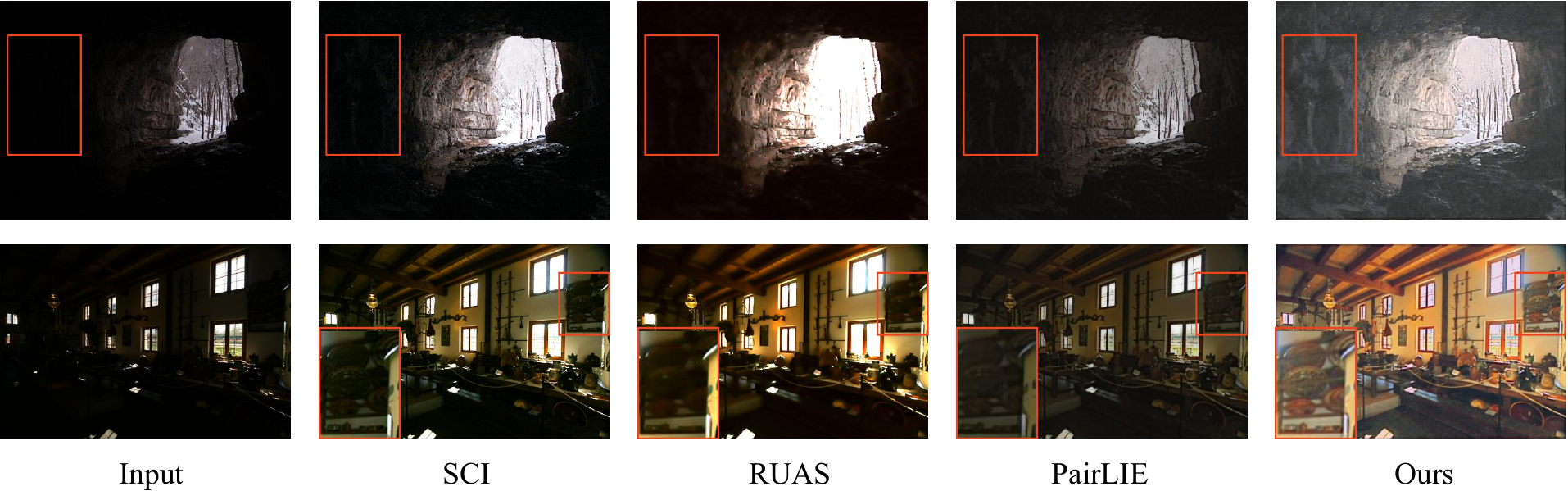}
    \caption{Additional qualitative results in edge situation of dataset MEF.}
    \label{fig:edge}
\end{figure}
\begin{figure}[h]
    \centering
    \includegraphics[width=\linewidth]{ICLR 2025 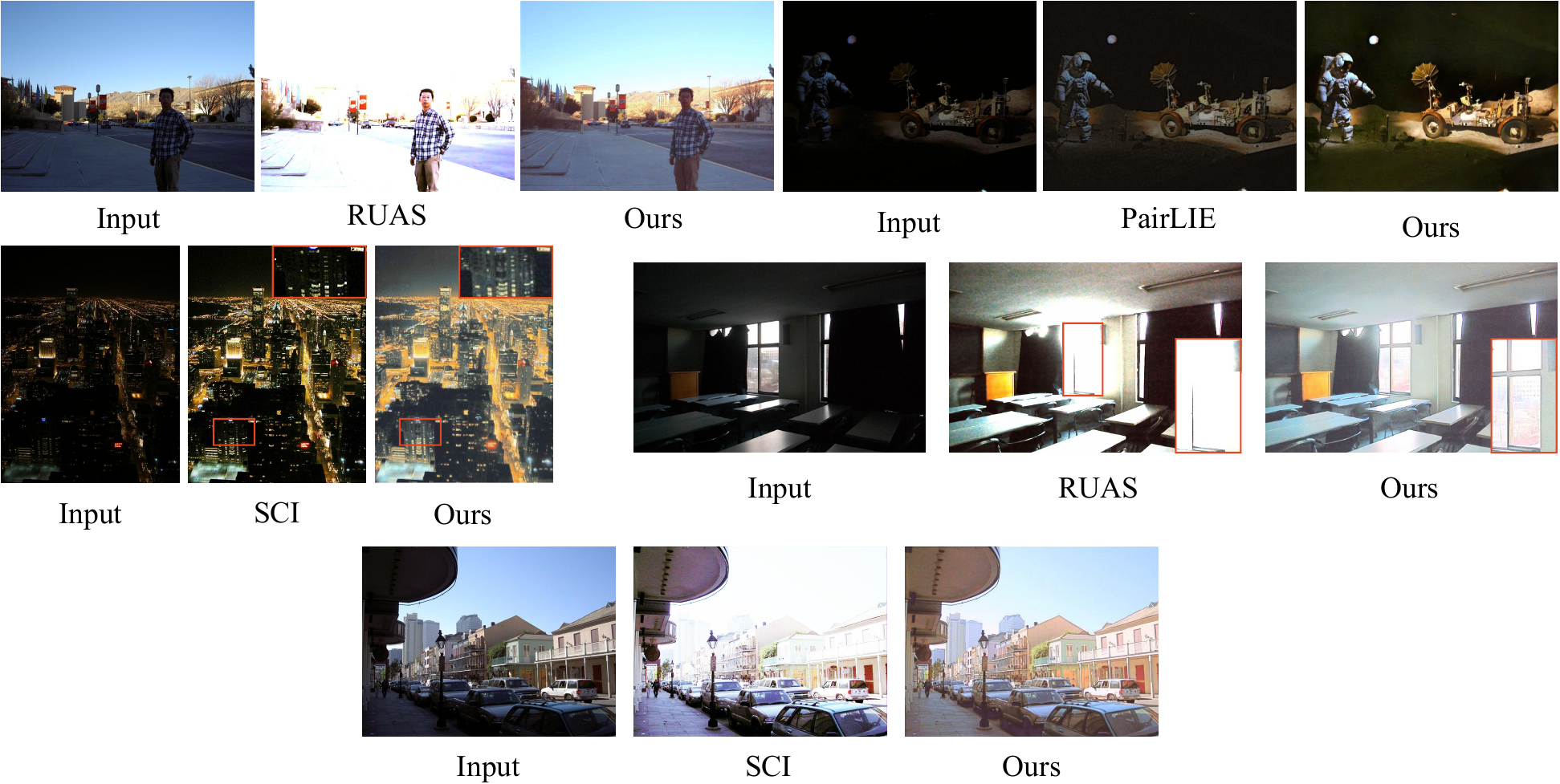}
    \caption{Additional qualitative results in edge situation of dataset DICM.}
    \label{fig:edge}
\end{figure}
\begin{figure}[h]
    \centering
    \includegraphics[width=\linewidth]{ICLR 2025 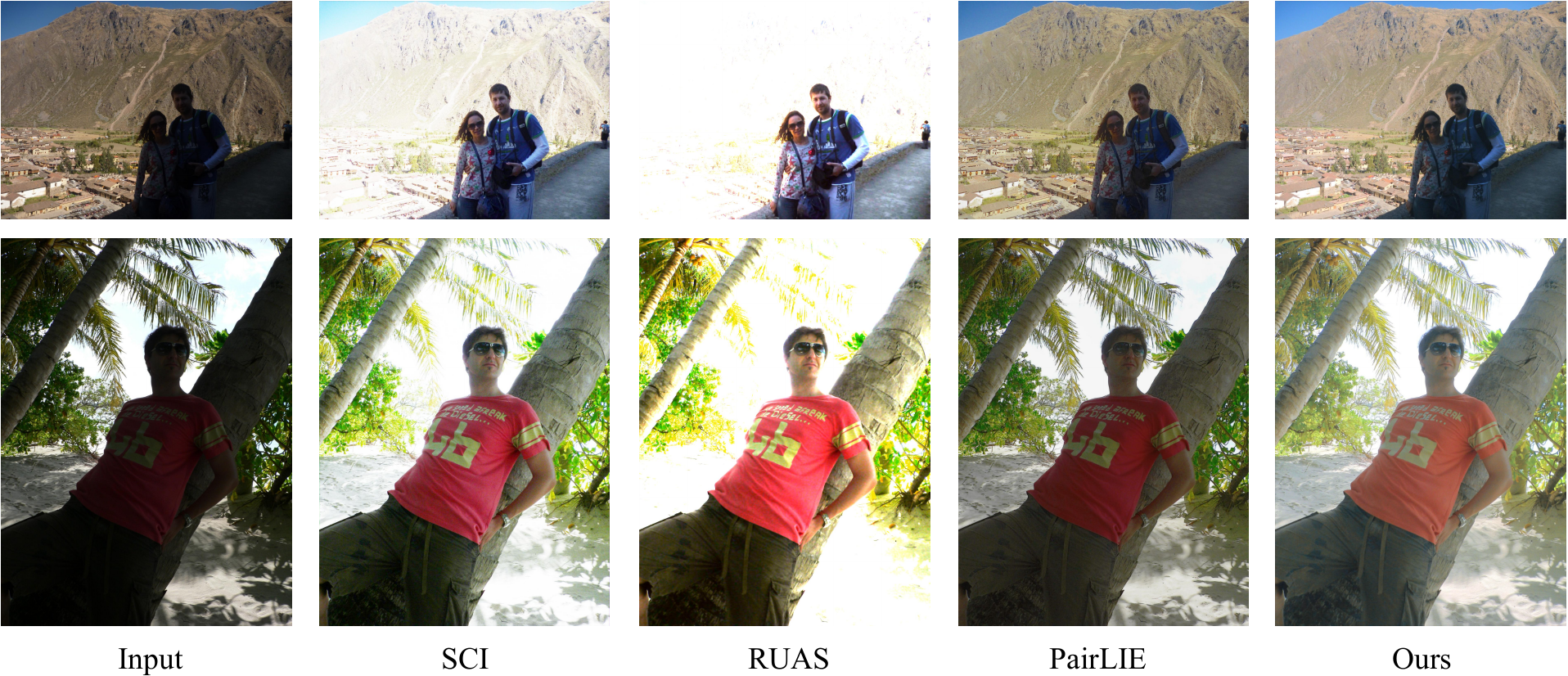}
    \caption{Additional qualitative results in edge situation of dataset VV.}
    \label{fig:edge}
\end{figure}

\bibliography{iclr2025_conference}
\bibliographystyle{iclr2025_conference}